\documentclass[accepted]{uai2026} 
                        
\usepackage[american]{babel}

\usepackage{natbib} 
\usepackage{mathtools} 
\usepackage{booktabs} 
\usepackage{tikz} 
\usepackage{cleveref}
\usepackage{bm}
\usepackage{graphicx}

\usepackage{notation}
\usepackage{amsfonts}
\usepackage{amsmath}
\usepackage{booktabs}
\usepackage{multirow}
\usepackage{researchpack}
\usepackage{algorithm}
\usepackage{algpseudocode}

\newcommand{\E}{\ensuremath{\mathbb{E}}}
\newcommand{\Cov}{\text{Cov}}

\newcommand{\varu}{\ensuremath{\boldsymbol{u}}}

\Crefname{thm}{Theorem}{theorems}
\Crefname{prop}{Proposition}{propositions}

\title{Zero-Variance Gradients for Variational Autoencoders}

\author[1]{Zilei Shao}
\author[2]{Anji Liu}
\author[1]{Guy Van den Broeck}
\affil[1]{%
    University of California, Los Angeles
}
\affil[2]{%
    School of Computing, National University of Singapore
}

\begin{document}
\maketitle

\begin{figure}[htbp]
    \centering
    \includegraphics[width=0.69\linewidth]{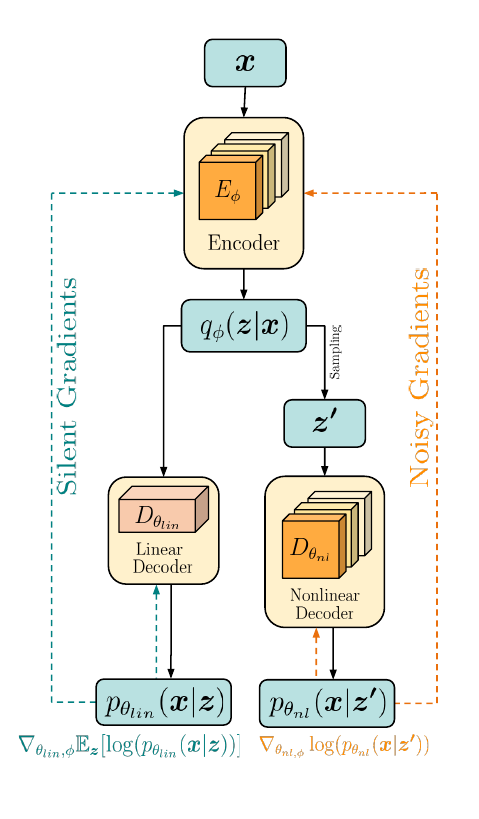}
    \caption{\textbf{Illustration of the use of Silent Gradients in training VAEs.} The encoder ($E_\phi$) takes input $\x$ and infers a latent distribution $q_\phi(\z|\x)$. These parameters are fed directly to the linear decoder ($D_{lin}$), which computes the analytical reconstruction log-likelihood, yielding a noise-free (Silent) gradient (dashed teal arrow) used to train the encoder. In parallel, samples $\z'$ are drawn from the latent distribution and fed to the nonlinear decoder ($D_{nl}$), which produces a standard, sample-based loss, resulting in a noisy gradient (dashed orange arrow). The solid black arrows represent the forward pass, while the dashed teal and orange arrows indicate the flow of gradients. During training, we can choose to train the encoder solely with the Silent Gradients or combine it with the noisy gradient using an annealing schedule. At inference time, only the trained encoder $E_\phi$ and nonlinear decoder $D_{nl}$ are used.}
    \label{fig:training_pipeline}
\end{figure}

\begin{abstract}
Training deep generative models like Variational Autoencoders (VAEs) requires propagating gradients through stochastic latent variables, which introduces estimation variance that can slow convergence and degrade performance. In this paper, we explore an orthogonal direction, which we call \textbf{Silent Gradients}. Instead of designing improved stochastic estimators, we show that by restricting the decoder architecture in specific ways, the expected ELBO can be computed analytically. This yields gradients with zero estimation variance as we can directly compute the evidence lower-bound without resorting to Monte Carlo samples of the latent variables. We first provide a theoretical analysis in a controlled setting with a linear decoder and demonstrate improved optimization compared to standard estimators. To extend this idea to expressive nonlinear decoders, we introduce a training paradigm that uses the analytic gradient to guide early encoder learning before annealing to a standard stochastic estimator. Across multiple datasets, our approach consistently improves established baselines, including reparameterization, Gumbel-Softmax, and REINFORCE. These results suggest that architectural choices enabling analytic expectation computation can significantly stabilize the training of generative models with stochastic components.
\end{abstract}

\section{Introduction}
Training of neural networks with stochastic components, such as the sampling of latent variables in generative models, often suffers from high variance in gradient estimates. This variance can impede the optimization process, leading to slower convergence and suboptimal model performance. 
In Variational Autoencoders (VAEs) \citep{vae, pmlr-v32-rezende14}, for instance, gradients must be propagated through a stochastic sampling layer. This has led to the development of several estimation techniques. For continuous latent spaces, the reparameterization trick \citep{vae} is commonly used. For discrete spaces, common approaches include the REINFORCE algorithm \citep{REINFORCE} and the Gumbel-Softmax trick \citep{gumbelmax, jang2017categoricalreparameterizationgumbelsoftmax}. However, all of these sample-based techniques introduce estimation variance, and in this paper we show that this variance hinders the optimization even in a simple, controlled setting.

In this paper, we propose a fundamentally different perspective on gradient estimation for VAEs. Given the estimation variance introduced by these stochastic gradient estimators, we argue for a different paradigm. Instead of developing more sophisticated techniques to \textit{estimate} the gradient of an expectation, we explore the possibility of first efficiently\footnote{In time linear in the number of latent dimensions.} computing the expectation itself in closed form, and then differentiating the resulting analytical expression. This path, when available, yields a gradient that is computed exactly and therefore has \textit{zero variance} by definition, in terms of the latent variables. Importantly, the zero-variance property specifically refers to the elimination of variance induced by latent variable sampling. Standard mini-batch stochasticity remains under stochastic optimization, and is orthogonal to our contribution. In practice, we find that the estimation variance can dominate the overall gradient noise, even under relatively small batch sizes where mini-batch variance would typically be expected to be substantial.

The feasibility of this approach hinges on the decoder architecture. While it is well-known that for a linear function, the expectation of its output can be computed exactly by linearity of expectation, this does not trivially extend to the full reconstruction log-likelihood. We first show that for a Gaussian likelihood with a fixed variance, the expected loss can still be computed in closed form, as a function of the latent distribution rather than the sampled latent variables. We then empirically demonstrate that using this analytic gradient leads to superior performance and faster convergence compared to standard stochastic estimators in this setting. 
Furthermore, we extend this technique to a more expressive setting where the output variance is also a learnable function of the latent distribution, again providing a zero-variance gradient. 
This analytic gradient component can then boost the performance of existing stochastic gradient estimators.

Finally, to generalize our method for more complex and practical settings, we introduce a novel training paradigm, depicted in \Cref{fig:training_pipeline}, that combines our analytic gradient component with standard, expressive nonlinear decoders. By using Silent Gradients to guide the initial training of the encoder before annealing to a conventional estimator, our technique serves as a powerful variance reduction tool that consistently improves the performance of established methods.
Our experimental results on the MNIST \citep{mnist}, ImageNet \citep{ImageNet}, and CIFAR-10 \citep{CIFAR10} datasets demonstrate a significant and consistent improvement in model performance. This shows that architectural choices that provide exact gradients are a powerful and general strategy for improving the training dynamics of models with stochastic layers.

\section{Background}

Variational Autoencoders (VAEs) \citep{vae} are a class of generative models for learning the probability distribution $p(\x)$ that underlies a dataset. VAEs introduce a set of latent variables $\z$ that are assumed to generate the observed data~$\x$. The model consists of two components: a prior distribution over the latent space $p(\z)$ and a conditional likelihood distribution $p_\theta(\x|\z)$, defined by the decoder $D_\theta$. The marginal likelihood of the model given the data then is $p_\theta(\x)=\mathbb{E}_{\z\sim p(\z)}[p_\theta(\x|\z)]$.

Since direct maximization of this likelihood is generally intractable, VAEs introduce a variational approximation $q_\phi(\z|\x)$ to the true posterior $p_\theta(\z| \x)$, called the encoder $E_\phi$, parameterized by $\phi$. Instead of maximizing the log-likelihood directly, one maximizes the Evidence Lower Bound (ELBO) \citep{ELBO}:
\begin{align}
\label{eq:ELBO}
    \mathbb{E}_{\z\sim q_\phi(\z | \x)}[\log p_\theta(\x | \z)] - D_{\mathit{KL}}(q_\phi(\z|\x)\,||\,p(\z)).
\end{align}

The first term is the expected reconstruction log-likelihood, which encourages the decoder to reconstruct the input $\x$ from its latent representation $\z$. The second term is the Kullback-Leibler (KL) divergence, which forces the approximate posterior $q_\phi(\z|\x)$ to be close to the prior $p(\z)$. Maximizing the ELBO provides a framework to jointly train the encoder and decoder parameters, $\phi$ and $\theta$, respectively.

\begin{table*}[ht]
\centering
\caption{\textbf{Decomposition of gradient variance into mini-batch variance (Batch Var) and estimator variance (Est Var).} 
The column ``Est \%'' reports the percentage of total variance due to latent sampling. 
Variance is computed using 50 mini-batches and 200 latent samples per example. 
Hyperparameters and training settings are consistent with those used in the experiments reported in \Cref{tab:mnist_results}.}
\label{tab:log_variance}

\resizebox{\linewidth}{!}{%
\begin{tabular}{llccccccccc}
\toprule
& & \multicolumn{3}{c}{Epoch 1} & \multicolumn{3}{c}{Epoch 50} & \multicolumn{3}{c}{Epoch 250} \\
\cmidrule(lr){3-5}\cmidrule(lr){6-8}\cmidrule(lr){9-11}
& Method 
& Batch Var & Est Var & Est \%
& Batch Var & Est Var & Est \%
& Batch Var & Est Var & Est \% \\
\midrule
\multirow{2}{*}{Continuous}
& Silent Gradients
& $1.49{\times}10^{4}$ & $0$ & $0.00$
& $5.28{\times}10^{2}$ & $0$ & $0.00$
& $4.53{\times}10^{1}$ & $0$ & $0.00$ \\
& Reparameterization
& $6.97{\times}10^{4}$ & $3.26{\times}10^{5}$ & $82.36$
& $2.96{\times}10^{3}$ & $3.28{\times}10^{4}$ & $91.72$
& $6.52{\times}10^{2}$ & $1.58{\times}10^{4}$ & $96.04$ \\
\midrule
\multirow{3}{*}{Discrete}
& Silent Gradients
& $2.18{\times}10^{5}$ & $0$ & $0.00$
& $4.53{\times}10^{6}$ & $0$ & $0.00$
& $1.12{\times}10^{5}$ & $0$ & $0.00$ \\
& Gumbel-Softmax
& $6.29{\times}10^{4}$ & $1.85{\times}10^{5}$ & $74.64$
& $2.12{\times}10^{4}$ & $2.32{\times}10^{5}$ & $91.63$
& $1.20{\times}10^{4}$ & $2.00{\times}10^{5}$ & $94.34$ \\
& REINFORCE
& $8.22{\times}10^{4}$ & $2.07{\times}10^{7}$ & $99.60$
& $4.86{\times}10^{5}$ & $8.30{\times}10^{7}$ & $99.42$
& $2.33{\times}10^{5}$ & $3.29{\times}10^{7}$ & $99.30$ \\
\bottomrule
\end{tabular}%
}
\end{table*}

\section{Exact ELBO with Linear Decoder}
\label{sec:fix-var-linear-dec}

While the KL divergence term is often analytically tractable by employing the mean-field assumption, in which the approximate posterior factorizes across latent dimensions: $q(\z|\x) := \prod_{i} q(z_i | \x)$, the reconstruction term, $\mathbb{E}_{\z\sim q_\phi(\z|\x)}[\log p_\theta(\x|\z)]$, remains intractable to compute. This difficulty comes from the complexity of the decoder function $p_\theta(\x|\z)$, which is typically a deep neural network. Consequently, estimating the gradient of this reconstruction term with respect to the encoder parameters $\phi$ poses a challenge since the gradient operator $\nabla_\phi$ cannot be passed inside an expectation that depends on those same parameters.

To resolve this problem, various techniques have been introduced \citep{REINFORCE, jang2017categoricalreparameterizationgumbelsoftmax, gumbelmax, vae}. However, we show that even widely-used estimators such as the reparameterization trick still have significant gradient variance. Specifically, we decompose the gradient variance of the ELBO into two components: (i) mini-batch variance, which arises from sampling different data batches and is shared by all methods, and (ii) estimator variance, which arises from sampling latent variables $z$ for a fixed batch. We measure both components on MNIST \citep{mnist} at different training stages (epochs 1, 50, and 250).

As shown in \Cref{tab:log_variance}, the overall gradient variance of standard stochastic estimators is substantial, and a large fraction of it attributes to latent sampling. In many cases, the variance brought by the estimator dominates the total variance, which highlights that the primary source of avoidable noise is the estimation variance arising from stochastic latent sampling. This naturally raises the question: how much performance is lost to this estimation variance, and what could be gained if we were able to compute the exact ELBO and its gradients? This motivates our exploration of an analytical approach. 

\subsection{Analytic gradient}
As established, the intractability of the reconstruction term comes from the complexity of the decoder $p_\theta(\x|\z)$, which is typically a deep neural network. This motivates an investigation into specific architectural choices where this analytical bottleneck can be resolved. 
We show that for a specific model structure, the expectation in the reconstruction term can be computed exactly (i.e., the first term in Eq.~\ref{eq:ELBO}), bypassing the need for stochastic estimation entirely. Let the data $\x$ and latent variable $\z$ be column vectors of dimensions $k$ and $d$, respectively. We consider a generative likelihood $p_\theta(\x|\z)$ that is a Gaussian distribution with a mean produced by a linear decoder and a fixed, scalar variance $\sigma^2$:
\begin{align}
    p_\theta(\x|\z) = \mathcal{N}(\x; \W_\mu \z, \sigma^2 I).
    \label{eq:fixed-var-linear-dec}
\end{align}
\noindent where the decoder is parameterized by the weight matrix $\W_\mu \in \mathbb{R}^{k\times d}$. For simplicity, we will denote the expectation $\mathbb{E}_{\bm{z}\sim q_\phi(\bm{z} | \bm{x})}$ as $\E$ where the context is clear.

Although this linear setup may seem restrictive, we will show in later sections that this technique forms the basis of a general method applicable to any VAEs. For now, we proceed by substituting this linear decoder structure into the ELBO reconstruction term:
\begin{align}
    \mathbb{E}[\log p_\theta(\bm{x} | \bm{z})] = -\frac{1}{2\sigma^2} \mathbb{E}[||\bm{x} - \W_\mu \bm{z}||^2_2] - \frac{1}{2} \log(2\pi\sigma^2).
    \label{eq:gauss-recon}
\end{align}
To compute this term exactly, all we need is to find an analytical form for the expectation $\mathbb{E}[||\bm{x} - \W_\mu \bm{z}||^2_2]$. We begin by expanding the squared $L_2$ norm:
\begin{align*}
    \mathbb{E}[||\x - \W_\mu \z||^2_2] &= \mathbb{E}[(\x - \W_\mu \z)^T(\bm{x}-\W_\mu \z)], \\
    &=\mathbb{E}[||\x||^2_2 - 2\x^T \W_\mu \z + ||\W_\mu \z||^2_2].
\end{align*}
By linearity of expectation, and because $\x$ and $\W_\mu$ are constants with respect to the expectation over $\bm{z}$, we simplify:
\begin{align*}
    \mathbb{E}[||\x - \W_\mu \z||^2_2] &=||\x||^2_2 - 2\x^T \W_\mu \mathbb{E}[\z] + \E[||\W_\mu \z||^2_2].
\end{align*}
The challenge lies in resolving the third term $\mathbb{E}[||\W_\mu \z||^2_2]$ since a naive computation of this expectation would take time $\bigO (k^2 d)$, which is quadratic w.r.t. the number of $\X$ variables. However, we can reduce the complexity by exploiting the fact that different variables in $\z$ are mutually independent due to the mean-field assumption.

We begin by expanding the quadratic term into a double summation $\mathbb{E}[||\W_\mu \z||^2_2] = \sum_i \sum_j \w_{\mu,i}^T\w_{\mu,j} \mathbb{E}[z_i z_j]$, where $\w_{\mu,i}$ is the $i$th column of $\W_\mu$. Using the identity $\mathbb{E}[z_iz_j] = \mathbb{E}[z_i]\mathbb{E}[z_j] + \text{Cov}(z_i, z_j)$, we split this summation into two parts:
\begin{equation*}
\begin{split}
    \sum_{i,j} \w_{\mu,i} ^T\w_{\mu,j} \mathbb{E}[z_i] \mathbb{E}[z_j] + \sum_{i,j} \w_{\mu,i} ^T\w_{\mu,j} \,\text{Cov}(z_i, z_j).
\end{split}
\end{equation*}
\noindent The first term, containing the expectations, can be factored into the square of a sum: $\left(\sum_i \w_{\mu,i}\mathbb{E}[z_i]\right)^2$, which can be computed in $\bigO(kd)$ time. The second covariance term can be simplified due to the independence of the latent variables. The covariance is zero for all $z_i, z_j$ ($i\neq j$) pairs, zeroing out all cross-terms. This collapses the double summation into a single sum over the variances $\text{Var}(z_i)$. The final analytical expression is the sum of these two simplified components:\footnote{The detailed step-by-step derivation is in Appendix \ref{appendix:squared_term}.}
\begin{equation}
\label{eq:sqaured_term}
     \mathbb{E}[||\W_\mu \z||^2_2]
    = ||\sum_i \w_{\mu,i}\mathbb{E}[z_i]||_2^2  + \sum_i||\w_{\mu,i}||_2^2 \text{Var}(z_i).
\end{equation}
Therefore, the expected squared error becomes:
\begin{equation*}
\begin{split}
    &\mathbb{E}[||\x - \W_\mu \bm{z}||_2^2] 
    = ||\x||^2_2 - 2\x^T \left(\sum_i\w_{\mu,i} \mathbb{E}[z_i]\right) \\
    &\quad\qquad \; + \left|\left|\sum_i \w_{\mu,i} \mathbb{E}[z_i]\right|\right|_2^2 + \sum_i ||\w_{\mu,i}||_2^2 \text{Var}(z_i).
\end{split}
\end{equation*}

Finally, we substitute this closed-form expression back into the expected reconstruction log-likelihood $\mathbb{E}[\log p_\theta(\x | \z)]$ and now this is fully analytical. The expectation over the latent variable $\bm{z}$ has been entirely eliminated, and the reconstruction loss now depends only on the mean $(\mathbb{E}[z_i])$ and variance $\text{Var}(z_i)$ of the latent distribution. This allows for direct and analytic gradient computation with respect to the encoder parameters.

\subsection{Do Analytic Gradients help?}
\label{sec:fixed_var_experiments}
As we show that the ELBO (and its gradients) can be computed exactly in an idealized setting with a linear decoder, we now investigate the practical impact of this zero-variance gradient. We conduct a controlled experiment on the MNIST dataset \citep{mnist} to quantify the performance gains from eliminating gradient estimation noise. This experiment uses the same simple VAE with a linear decoder and fixed output variance ($\sigma^2 = 0.01$) (cf. Eq.~(\ref{eq:fixed-var-linear-dec})) as our our gradient variance analysis in \Cref{tab:log_variance}, a setting designed to isolate the estimator's impact rather than to achieve state-of-the-art results. To ensure a fair comparison, we perform a separate hyperparameter search for each method. By epoch 500, all models have converged, and we report all final metrics at this point. All further details regarding the model architecture and hyperparameters can be found in Appendix \ref{appendix:experiment_details}.
We benchmark our method, Silent Gradients, against the reparameterization trick in the continuous latent space and against the Gumbel-Softmax estimator and the REINFORCE algorithm in the discrete space. Model performance is evaluated using Bits Per Dimension (BPD) and Mean Squared Error (MSE).
 
\begin{table}[tb]
\centering
\caption{\textbf{Performance comparison on MNIST using a linear decoder with a latent dimension of 200 under fixed variance $\sigma^2 = 0.01$.} The best BPD and MSE values for the continuous and discrete latent space are in bold, respectively. Silent Gradients consistently outperforms stochastic gradient estimators in terms of BPD and MSE.}
\label{tab:mnist_results}
\resizebox{\linewidth}{!}{%
\begin{tabular}{llcc}
\toprule
& \multirow{2}{*}[-0.1em]{Method} & \multicolumn{2}{c}{MNIST} \\
& & BPD ($\downarrow$) & MSE ($\downarrow$)\\
\midrule
\multirow{2}{*}{Continuous} & Silent Gradients & \textbf{6.718} & \textbf{3.011} \\
& Reparameterization & 6.722 & 3.059 \\
\midrule
\multirow{3}{*}{Discrete} & Silent Gradients & \textbf{6.900} & \textbf{6.103} \\
& Gumbel-Softmax & 6.990 & 7.670 \\
& REINFORCE & 7.208 & 9.289 \\
\bottomrule
\end{tabular}
}
\end{table}

The results, summarized in \Cref{tab:mnist_results}, demonstrate the consistent advantage of our method. In the discrete latent space setting, our method achieves a substantially lower BPD than the corresponding baselines. In the continuous case, while the final BPD scores are comparable, our method demonstrates significantly faster convergence; Silent Gradients reaches 6.73 BPD in just 45 epochs, a milestone that standard reparameterization requires 90 epochs to achieve.

Besides BPD scores, the low MSE of our method confirms high-fidelity image reconstruction, indicating that the reported BPD is not limited by poor reconstruction quality but is instead constrained by the fixed-variance assumption. The sharp reconstructions in Appendix \ref{appendix:visualized_output} visually corroborate this conclusion.

\section{More Expressive Decoders}
We have shown that Silent Gradients offers a significant performance boost when the analytic gradient of the ELBO can be tractably computed. However, it is still unclear how to apply our method to VAEs with more general decoders. We address this in two steps: first, we demonstrate how to generalize the linear Gaussian decoder setting to make the variance a learnable parameter. Next, we show that this tractable linear component can be integrated with any existing VAE to guide encoder learning.

\subsection{Linear Decoders with Learnable Variance}
A key limitation of the fixed-variance Gaussian decoder introduced in the previous section is its inability to dynamically adjust confidence across different variables, resulting in significant performance degradation. This motivates generalizing our approach to allow variance to be a learnable, data-dependent function of the latent variable $\z$. That is, given latents $\z$, we predict both the mean $\mu (\z)$ and the variance $\sigma^2 (\z)$ of the Gaussian distribution.

Under this parameterization, the first term of the expected reconstruction log-likelihood (\ie Eq.~(\ref{eq:gauss-recon})) is generalized to $\E[\frac{(\x-\mu(\z))^2}{2\sigma^2(\z)}]$. Computing the expectation of a reciprocal is \#P-hard for simple function classes including multilinear polynomials \citep{atlas}. Furthermore, to ensure the expression is well-defined, $\sigma^2 (\mathbf{z})$ must be strictly positive. While this can be enforced through techniques such as lower-bounding the variance by clipping, these approaches introduce discontinuities that complicate the analytical computation of the expectation and may hinder stable optimization. In addition, the second term in the reconstruction log-likelihood, $\frac{1}{2}\log(2\pi \sigma^2(\z))$, that involves the expectation of a logarithm, is also computationally hard \citep{atlas}. 

To sidestep these challenges, we propose to represent the scale of the Gaussian distribution by the reciprocal of the standard deviation.\footnote{This parametrization aligns with the classical notion of precision, as originally defined by \citet{gauss1877theoria}; today the term precision is also used to denote the reciprocal of the variance.} Formally, this quantity is called precision and is defined as $\alpha (\z) = 1 / \sigma (\z)$. 

Following \Cref{sec:fix-var-linear-dec}, we define both the mean $\mu(\z)$ and the precision $\alpha (\z)$ as linear functions of the latent variable $\z$: 
\begin{align*}
    \mu(\z) = \W_\mu \z,\quad  \alpha(\z) = \W_\alpha \z,
\end{align*}
\noindent which gives the model flexibility to assign pixel-wise uncertainty. Following \cref{eq:fixed-var-linear-dec}, we define the generative likelihood as a Gaussian distribution where the mean is given by $\mu(\z)$ and the variance is the element-wise inverse square of $\alpha(\z)$:
\begin{align}
    p_\theta(\x|\z) = \mathcal{N}\left(\x; \mu(\z), \text{diag}\left(\frac{1}{\alpha(\z)^2}\right)\right).
\end{align}
By the definition of the precision, the expected reconstruction log-likelihood from the ELBO becomes:
\begin{equation}
\label{eq:log_likelihood}
\begin{split}
    \mathbb{E}[\log p_\theta(x | z)]
    &= -\frac{1}{2}\E\left[ ||(\x -\mu(\z))\odot \alpha(\z)||_2^2\right] \\
    & \qquad  +\frac{1}{2}\E[\log(\alpha(\z)^2)]- \frac{1}{2} \log(2\pi).
\end{split}
\end{equation}

The exact computation of both the first term and the second term is non-trivial. The first term involves an expectation of products of correlated functions of $\z$. The second term is hard since $\E[\log(\alpha(\z))] \neq \log(\E[\alpha(\z)])$.

We expand the first expectation term as follows:
\begin{equation}
\begin{split}
    & \mathbb{E}\left[ || (\x -\mu(\z))^T \alpha(\z) ||_2^2\right] = \\
    & \qquad \mathbf{1}^T \big ( ||\x||_2^2 \, \E \left[||\W_\alpha\z||^2_2 \right] - 2 \x\odot \E\left[\W_\mu\z||\W_\alpha\z||_2^2\right] \\\nonumber
    & \qquad \qquad  +\E\left[||\W_\mu\z||^2_2||\W_\alpha\z||_2^2\right] \big ).
\end{split}
\end{equation}

The first term in this expansion, $\E \left[||\W_\alpha\z||^2_2 \right]$, is identical in form to the quadratic term derived in the fixed-variance setting (i.e. Eq.~\ref{eq:sqaured_term}). As we showed previously, it can be computed analytically, depending on only the mean and variance of latent distribution. 

For the two remaining terms, $\E [\W_\mu \z||\W_\alpha\z||^2_2]$, and $\E[||\W_\mu\z||^2_2||\W_\alpha\z||_2^2]$, we begin the derivation by separating the terms into their expected values and covariances:
\begin{align*}
    \E\left[\W_\mu\z||\W_\alpha\z||_2^2\right] &= \mathbb{E}[\W_\mu \z]\mathbb{E}[||\W_\alpha \z||_2^2] \notag\\
    &\quad+ \text{Cov}(\W_\mu \z, ||\W_\alpha \z||_2^2);
\end{align*}
\begin{align*}
        \E\left[||\W_\mu\z||^2_2||\W_\alpha\z||_2^2\right] &= \mathbb{E}[||\W_\mu \z||_2^2]\mathbb{E}[||\W_\alpha \z||_2^2] \notag\\
        &\quad +\text{Cov}(||\W_\mu z||_2^2, ||\W_\alpha z||_2^2).
\end{align*}
The challenge, therefore, lies in deriving the two covariance terms. Following the principles for computing the covariance of products of random variables by \citet{covariance}, these terms can be decomposed into functions of central moments of the individual latent variables $z_i$. This makes the tractability of the entire expression dependent on whether these underlying central moments can be computed in closed form, as shown below.

\begin{prop}[Tractable Central Moments]
\label{prop:1}
    Let $\z\in \mathbb{R}^d$ be a random vector with independent components $z_i$. The first four central moments of each component, $\E[\tilde{z_i}]:=\E[(z_i-\E[z_i])^k]$ for $k\in\{1,2,3,4\}$, can be computed in closed form of the parameters of its distribution if $z_i$ follows:

     1. A Gaussian distribution, $z_i\sim \ensuremath{\mathcal{N}}(\mu_i, \sigma_i^2)$. 
     
     2. A Bernoulli distribution, $z_i\sim \text{Bern}(p_i)$.
\end{prop}

\begin{proof}[Derivation Sketch] 
The proof follows from the definitions of the moment-generating functions for each distribution. For a Gaussian variable, the central moments can be derived to be simple functions of its variance $\sigma_i^2$ \citep{winkelbauer2014momentsabsolutemomentsnormal}. For a Bernoulli variable with probability $p_i$, the raw moments $\E[z_i^k]$ are trivial to compute, and the central moment of order $k$ is given by $(1-p_i)(-p_i)^k + p_i(1-p_i)^k$, resulting in polynomials of $p_i$. The full derivations are provided in the Appendix \ref{appendix:derivation}.
\end{proof}

With tractability of the individual central moments established, we can now show how this allows for the analytical computation of the covariance terms.

\begin{thm}[Analytic Covariance of Linear Projections]
\label{thm:1}
    Let $\W_\mu \z$ and $\W_\alpha \z$ be two linear projections of a random vector $\z$ whose components $z_i$ are independent. The covariance terms $\text{Cov}(\W_\mu \z, ||\W_\alpha \z||^2)$ and $\text{Cov}(||\W_\mu \z||_2^2, ||\W_\alpha \z||_2^2)$ can be expressed as a linear combination of the first four central moments of the components $z_i$, respectively. The coefficients of this linear combination are polynomials in the entries of the matrices $\W_\mu$ and $\W_\alpha$.
\end{thm}

\begin{proof}[Proof Sketch] 
The proof relies on the formula for the covariance of products of random variables. The full derivation is in Appendix \ref{appendix:derivation}.

1. For the term $\text{Cov}(\W_\mu \z, ||\W_\alpha \z||_2^2)$: We decompose this covariance term into a function of third-order expectations, $\E[\tilde{z_i}\tilde{z_j}\tilde{z_k}]$, where $\tilde{\z} = \z - \E[\z]$. Due to the independence of the latent variables $z_i$, these expectations are non-zero only when all indices are identical ($i=j=k$). This simplifies the expression to a function of the second and third central moments of $z_i$. The resulting expression is a linear combination of the second and third central moments of $z_i$. Its coefficients are third-degree polynomials of the weight matrices $\W_\mu$ and $\W_\alpha$.

2. For the term $\text{Cov}(||\W_\mu \z||_2^2, ||\W_\alpha \z||_2^2)$: Similarly, this term can be decomposed into functions of fourth-order expectations, $\E[\tilde{z_i}\tilde{z_j}\tilde{z_k}\tilde{z_l}]$. Under the independence assumption, these complex expectations simplify into a linear combination of the second, the third, and the fourth central moments. The coefficients are fourth-degree polynomials of the weight matrices.
\end{proof}

While the covariance terms can be computed analytically, the full log-likelihood function as in \Cref{eq:log_likelihood} still contains the intractable logarithmic term $\E[\log(\W_\alpha \z)]$, To ensure the argument of the logarithm is non-negative and to maintain a tractable, zero-variance objective, we approximate the term $\E[\log(||\W_\alpha \z||_2^2)]$ using a second order Taylor Expansion around the mean of $||\W_\alpha\z||_2^2$ \citep{logtermtaylorexpansion}:
\begin{align*}
    \mathbb{E}[\log(||\W_\alpha \z||_2^2)] \approx \log(\E[||\W_\alpha \z||_2^2)] \!-\! \frac{\text{Var}[||\W_\alpha \z||_2^2]}{2(\mathbb{E}[||\W_\alpha \z)^2]||_2^2}.
\end{align*}
To assess the second-order Taylor approximation of the logarithmic term, we conducted a controlled experiment in which we brute-force enumerated the discrete latent distribution to obtain the exact gradient as ground truth. The bias introduced by this approximation is negligible in practice. With learnable variance, the $L_2$ deviation of Silent Gradient (212.9) is substantially smaller than that of Gumbel-Softmax (6.1k) and even the 100{,}000-sample estimation error of REINFORCE (593.1), showing that the approximation error is minor relative to stochastic estimation noise.

By combining the exact computations for the covariance terms with this approximation, the expected reconstruction log-likelihood can be expressed in an analytical solution:
\begin{align}
\label{eq:expected_log_likelihood}
        &\quad \mathbb{E}[\log p_\theta(\bm{x}|\bm{z})] \notag\\
         &\approx -\frac{1}{2}\Bigl(||\x||_2^2\mathbb{E}[||\W_\alpha \z||_2^2] - 2\x^T\bigl(\W_\mu \mathbb{E}[\z]\mathbb{E}[||\W_\alpha \z||^2] 
        \notag\\
        &\quad
        + \text{Cov}(\W_\mu \z, ||\W_\alpha \z||_2^2)\bigr) + \mathbb{E}[||\W_\mu \z||_2^2]\mathbb{E}[||\W_\alpha \z||_2^2] 
        \notag\\
        &\quad
        + \Cov(||\W_\mu \z||_2^2, ||\W_\alpha \z||_2^2)\Bigr)- \frac{1}{2}\log(2\pi)
         \notag\\
        &\quad
        +\left( \frac{1}{2}\log(\mathbb{E}[||\W_\alpha \z||_2^2]) - \frac{\text{Var}[||\W_\alpha \z||_2^2]}{4(\mathbb{E}[||\W_\alpha \z||_2^2])^2} \right).
\end{align}

This expression now relies only on the tractable moments of the latent distribution and the decoder weights.

\subsection{Silent Gradients with General VAEs}
\begin{table*}[ht]
\centering
\caption{\textbf{Performance comparison in BPD ($\downarrow$) for models with learnable variance across datasets.} Results are reported as mean $\pm$ standard deviation over 3 random seeds. For each baseline, the lower mean BPD between ``Without SG'' and ``With SG'' is shown in bold. Silent Gradients (SG) consistently improves standard estimators across datasets.}
\label{tab:learnable_var_results}
\resizebox{\linewidth}{!}{%
\begin{tabular}{llcccccc}
\toprule
& \multirow{2}{*}[-0.3em]{Method} 
& \multicolumn{2}{c}{MNIST} 
& \multicolumn{2}{c}{ImageNet} 
& \multicolumn{2}{c}{CIFAR-10} \\
\cmidrule(lr){3-4} \cmidrule(lr){5-6} \cmidrule(lr){7-8}
& & Without SG & With SG 
& Without SG & With SG 
& Without SG & With SG \\
\midrule
\multirow{2}{*}{Continuous}
& None 
& -- 
& 2.36 $\pm$ 0.07 
& -- 
& 5.99 $\pm$ 0.01 
& -- 
& 5.78 $\pm$ 0.03 \\

& Reparameterization 
& 1.95 $\pm$ 0.04 
& \textbf{1.83} $\pm$ 0.03 
& 5.81 $\pm$ 0.02 
& \textbf{5.70} $\pm$ 0.01 
& 5.71 $\pm$ 0.01 
& \textbf{5.53} $\pm$ 0.01 \\
\midrule
\multirow{3}{*}{Discrete}
& None 
& -- 
& 2.74 $\pm$ 0.06 
& -- 
& 6.46 $\pm$ 0.01 
& -- 
& 6.72 $\pm$ 0.01 \\

& Gumbel-Softmax 
& 2.50 $\pm$ 0.02 
& \textbf{2.37} $\pm$ 0.04 
& 6.31 $\pm$ 0.01 
& \textbf{6.20} $\pm$ 0.01 
& 6.21 $\pm$ 0.01 
& \textbf{6.20} $\pm$ 0.02 \\

& REINFORCE 
& 2.99 $\pm$ 0.08 
& \textbf{2.93} $\pm$ 0.02 
& 6.88 $\pm$ 0.01 
& \textbf{6.78} $\pm$ 0.01 
& 6.74 $\pm$ 0.01 
& \textbf{6.67} $\pm$ 0.01 \\
\bottomrule
\end{tabular}
}
\end{table*}

While the preceding section demonstrates that analytical gradients are tractable for linear decoders with learnable variance, the expressive power of a purely linear model is limited. To handle more complex data distribution, we now introduce a training strategy that integrates the benefits of our tractable Silent Gradients with general, powerful nonlinear decoders.

Our approach uses a dual-decoder architecture consisting of a shared encoder, a linear decoder for computing the exact ELBO component and computing the exact Silent Gradients, and a parallel, more expressive nonlinear decoder for generating the final reconstructions. A visualization of this pipeline is presented in \Cref{fig:training_pipeline}. Although the linear and nonlinear decoders differ in expressive power, both are trained to reconstruct the same data distribution. As a result, their gradients with respect to the shared encoder are aligned toward improving reconstruction. In early training, the linear decoder provides a coarse but stable reconstruction signal, effectively sketching the data manifold and guiding the encoder toward a useful latent structure before the nonlinear decoder refines it.

\begin{algorithm}[t]
\caption{Training Dynamics for Integrating Silent Gradients}
\label{alg:training_dynamic}
\begin{algorithmic}[1]
\Require Encoder $E_\phi$, Linear Decoder for $\alpha(\z)$ $D_{lin, \alpha}$, Linear Decoder for $\mu(\z)$ $D_{lin, \mu}$, Nonlinear Decoder $D_{nl}$
\Require Training data $\mathcal{D}$, cut-off epoch $N_{{cutoff}}$, annealing rate $\lambda$

\For{$n_{epoch}$ = 1 to $N_{max}$}
    \If{$n_{epoch}$ == $N_{{cutoff}}$}
        \State Freeze parameters $\phi$ of the encoder $E_\phi$
    \EndIf
    \For{batch $\bm{x}$ in $\mathcal{D}$}
        \State $\bm{z}, \text{stats} \gets E_\phi(\bm{x})$
        \State $\mathcal{L}_{lin} \gets -D_{lin}(\text{stats}, \bm{x})$ \Comment{Analytical ELBO component \Cref{eq:expected_log_likelihood}}
        \State $\mathcal{L}_{nl} \gets -\log p_{nl}(\bm{x}|\bm{z})$ \Comment{Sampled reconstruction loss}

        \State $w_{lin} \gets \max(0, 1 - n_{epoch} \cdot \lambda)$
        \State $w_{nl} \gets 1 - w_{lin}$
        
        \State $\mathcal{L}_{recon} \gets w_{lin} \cdot \mathcal{L}_{lin} + w_{nl} \cdot \mathcal{L}_{nl}$
        \State $\mathcal{L}_{total} \gets \mathcal{L}_{recon} + D_{KL}$
        \State Take gradient step on $\mathcal{L}_{total}$ for all unfrozen parameters
    \EndFor
\EndFor
\end{algorithmic}
\end{algorithm}

The training follows a two-stage process. In the initial stage, the encoder and both decoders are trained, but the encoder parameters are updated only using the analytic gradients from the linear decoder. After a set number of epochs, we freeze the encoder's weights. In the second stage, only the nonlinear decoder continues to train, fine-tuning its parameters on the now fixed, well-structured latent space provided by the encoder. This configuration corresponds to the “None + With SG” setting in our experimental comparison, as shown in \Cref{tab:learnable_var_results} and \Cref{tab:kld_results}.

The Silent Gradients framework can be further extended to boost the performance of existing gradient estimators. Instead of having the encoder rely solely on the linear decoders' analytical gradient, we introduce a gradient annealing schedule. In this combined approach, the gradient signal sent to the encoder $E_\phi$ is a weighted average:
\begin{align}
    \nabla_{\phi, \text{total}} = w_{\text{lin}} \nabla_{\phi, \text{Silent}} + w_{\text{nl}} \nabla_{\phi, \text{Noisy}},
\end{align}
\noindent where $w_{\text{nl}} = 1-w_{\text{lin}}$, $\nabla_{\phi, \text{Silent}}$ is the analytical gradient from the linear decoder, and $\nabla_{\phi, \text{Noisy}}$ is the noisy gradient from the nonlinear decoder using stochastic estimators. The training begins with the weight of the Silent Gradients, $w_{lin}$, at 1.0 and the weight of the baseline estimator's gradient, $w_{nl}$, at 0.0. As training progresses, $w_{lin}$ is gradually annealed to 0 while $w_{nl}$ is increased to 1.0. This dynamic allows the encoder to first learns a representation guided by the noise-free, analytical signal before fine-tuning with the sample-based gradients from the full, expressive model. The complete training dynamic is detailed in \Cref{alg:training_dynamic}. This configuration corresponds to the “With SG” variants of the baseline estimators in our experimental comparison.

\begin{table*}[ht]
\centering
\caption{\textbf{KL Divergence (KLD) comparison for models with learnable variance.} The KLD for the combined method is in bold when it is higher than its corresponding baseline. The results consistently show a higher KLD when a baseline estimator is combined with our Silent Gradient technique, which suggests the encoder learns a more informative latent representation.}
\label{tab:kld_results}
\begin{tabular}{llcccccc}
\toprule
& \multirow{2}{*}[-0.3em]{Method} & \multicolumn{2}{c}{MNIST} & \multicolumn{2}{c}{ImageNet} & \multicolumn{2}{c}{CIFAR-10} \\
\cmidrule(lr){3-4} \cmidrule(lr){5-6} \cmidrule(lr){7-8}
& & Without SG & With SG & Without SG & With SG & Without SG & With SG \\
\midrule
\multirow{2}{*}{Continuous} & None & -- & 330.77 & -- & 478.43 & -- & 550.70 \\
& Reparameterization & 155.15 & \textbf{165.76} & 382.87 & \textbf{533.20} & 427.79 & \textbf{577.25} \\
\midrule
\multirow{3}{*}{Discrete} & None & -- & 91.82 & -- & 534.29 & -- & 243.69 \\
& Gumbel-Softmax & 94.64 & \textbf{96.24} & 368.02 & \textbf{404.57} & 446.75 & 442.58 \\
& REINFORCE & 109.67 & \textbf{128.55} & 294.88 & \textbf{381.27} & 303.33 & \textbf{367.10} \\
\bottomrule
\end{tabular}
\end{table*}

We benchmark both our standalone Silent Gradients method and the combined approach against baselines on MNIST, ImageNet, and CIFAR-10. All models were tuned for optimal hyperparameters and trained until convergence to ensure a fair comparison. Our experimental results, presented in \Cref{tab:learnable_var_results}, demonstrate two key findings. First, our Silent Gradients method consistently improves the performance of existing gradient estimators. In every case, combining a standard estimator with our technique results in a lower BPD score compared to the baseline alone across all tested datasets. This shows that our analytical gradient serves as a powerful and general-purpose training aid. Second, our method used as a standalone estimator is highly competitive, even outperforming the widely used REINFORCE estimator on both MNIST and ImageNet. Additional quantitative results and experimental details are provided in Appendix \ref{appendix:experiment_details}, with reconstruction visualizations shown in Appendix \ref{appendix:visualized_output}.

An analysis of the KL Divergence (KLD) offers an explanation for these performance gains. It is important to clarify that our objective is to maximize the ELBO, not to minimize the KL divergence term in isolation. A higher KL divergence indicates that the encoder is utilizing the latent space more effectively. In our experiments shown in \Cref{tab:kld_results}, models trained with Silent Gradients achieve both higher KL divergence and lower reconstruction loss, resulting in a strictly higher ELBO. This suggests that Silent Gradients mitigates posterior collapse, encouraging a more informative latent representation. We hypothesize this is because the zero-variance analytical gradient provides a cleaner, more stable training signal to the encoder than the noisy gradients from the standard stochastic estimators.

While our results are not state-of-the-art in absolute BPD, our goal is to isolate and evaluate the impact of gradient variance rather than optimize architecture or scale. To this end, we intentionally adopt a simple decoder setting that allows a controlled comparison across estimators. Within this controlled regime, Silent Gradients consistently achieves meaningful performance gains across all baselines, highlighting its effectiveness as a general optimization tool for generative models.

\section{Related Work}
\paragraph{VAEs.}

Variational Autoencoders (VAEs) are generative models that learn a latent representation of data through an encoder-decoder framework \citep{vae}. They can be categorized by their latent space: VAEs with continuous latent variables typically use Gaussian distributions and are widely applied to tasks like image modeling \citep{vae}, while VAEs with discrete latent spaces have become an active research area, as discrete representations can offer better interpretability and computational efficiency \citep{vqvae, jang2017categoricalreparameterizationgumbelsoftmax}. This line of work contains various architectures of the discrete latent space, such as the use of vector quantization in VQ-VAE \citep{vqvae} and relaxed Boltzmann priors in DAVE\# \citep{dvae2}.

\paragraph{Gradient Estimation Techniques.}
A key challenge in training VAEs is propagating gradients through stochastic sampling layers. In the continuous case, the reparameterization trick, which separates the stochasticity into a fixed noise source and a deterministic function, is widely used \citep{vae}. Although unbiased, reparameterization still introduces variance that impedes optimization.

In the discrete case, two main lines of techniques are used. The first is the use of the REINFORCE technique, or score function estimator, which provides a general and unbiased gradient estimate applicable to both discrete and continuous latent variables. It rewrites the gradient of the expectation as: $\nabla_\phi \E_{q_\phi(\z)} f(\z) = \E_{q_\phi(\z)}[f(\z)\nabla_\phi\log q_\phi(\z)]$ \citep{REINFORCE}. However, this estimator is often hindered by high variance, which has led to the development of variance reduction techniques such as control variates, \citep{REINFORCE_baseline, REINFORCE_sample_baseline}.

The second line of research strives to make discrete variables compatible with low-variance reparameterization trick. The straight-through (ST) estimator approximates the discrete sampling in the backward pass with a differentiable function, such as using the mean value for a Bernoulli variable \citep{straight_through}. Another approach is to relax discrete variables into a continuous distribution; the Concrete \citep{gumbelmax} or Gumbel-Softmax \citep{jang2017categoricalreparameterizationgumbelsoftmax} distribution, for instance, achieves this by adding Gumbel noise to the logits of a softmax function, enabling reparameterization. More recent techniques such as SIMPLE \citep{SIMPLE}, IndeCateR \citep{catlog}, and Implicit Maximum Likelihood Estimation (IMLE) \citep{IMLE} offer alternative strategies to derive low-variance gradient estimates for generative models with discrete latent variables.

\paragraph{Linear VAEs.} Linear VAEs are a cornerstone in various contexts. First, their analytical tractability makes them an ideal setting for theoretical investigation. For example, \citet{10.5555/3454287.3455131} used linear VAEs to show that posterior collapse can be an inherent issue of the marginal log-likelihood objective, not a problem caused by the ELBO approximation. Other work uses them to investigate the implicit bias of gradient descent, showing how training dynamics can recover the ground-truth data manifold \citep{koehler2022variational}. Additionally, linear decoders are also crucial in tasks such as learning sparse and interpretable features from complex data \citep{lu2025sparse,vafaii2024poisson}.
This broad utility motivates our method, which allows for analytic gradient estimation for any VAE with a linear decoder. Having demonstrated its effectiveness in image modeling, Silent Gradients could directly enhance these other applications.

\paragraph{Analytic ELBO.} \citet{10.5555/3454287.3455131} derive an analytical ELBO for linear VAEs under assumptions of a fixed scalar output variance, a Gaussian latent space and a linear encoder. In contrast, our method is more general, that supports a learnable variance for output Gaussian distribution, applies to any latent distribution with tractable central moments, and makes no assumptions about the encoder architecture.

\section{Conclusion}
In this work, we introduced Silent Gradients, a new perspective on training VAEs that targets the variance introduced by stochastic latent sampling. Instead of improving stochastic estimators, we leverage specific decoder architectures to analytically compute a zero-variance gradient signal. We provided theoretical derivation and empirically showed its effectiveness. Our method not only outperforms standard estimators in a controlled setting but also consistently improves their performance when integrated into general VAEs through a simple annealed training scheme.

Beyond linear decoders, our framework points to a broader direction: integrating tractable probabilistic models into deep generative architectures to enable exact or partially exact expectation computation. In particular, Silent Gradients can extend naturally to expressive families such as Probabilistic Circuits, which support exact probabilistic queries, model mixtures, and capture variable correlations. Such integrations may further reduce estimation noise while preserving expressive power.

\section{Acknowledgments}
This work was funded in part by the National University of Singapore under its Start-up Grant (Award No: SUG-251RES2505), the DARPA ANSR, CODORD, and SAFRON programs under awards FA8750-23-2-0004, HR00112590089, and HR00112530141, NSF grant IIS1943641, and gifts from Adobe Research, Cisco Research, Qualcomm, and Amazon. Approved for public release; distribution is unlimited. The authors would like to thank Benjie Wang for useful discussion on the proofs.

\bibliography{reference}

\newpage
\onecolumn
\setcounter{secnumdepth}{2} 
\appendix
\section{Derivation}
\label{appendix:derivation}

In this section, we provide the step-by-step derivation for \Cref{eq:sqaured_term}, \Cref{prop:1}, and \Cref{thm:1}.

\subsection{\texorpdfstring{$\E||\W_\mu \z||^2_2$}{E[||Wmu z||^2_2]}} 
\label{appendix:squared_term}
We wish to prove: 
\begin{equation}
     \mathbb{E}[||\W_\mu \z||^2_2]
    = ||\sum_i \w_{\mu,i}\mathbb{E}[z_i]||_2^2  + \sum_i||\w_{\mu,i}||_2^2\text{Var}(z_i).
\end{equation}

\noindent \textbf{Derivation.} To begin with, we shall expand $\E||\W_\mu \z||^2_2$ using summations:
\begin{align}
    \E[||\W_\mu \z||^2_2] &= \mathbb{E}\left[||\sum_i \w_{\mu,i} z_i||_2^2\right], \\
    &= \mathbb{E}\left[\sum_i \sum_j (\w_{\mu,i} z_i)^T(\w_{\mu,j} z_j)\right], \\
    &= \sum_i \sum_j \w_{\mu,i}^T\w_{\mu,j} \mathbb{E}[z_i z_j]. 
\end{align}
\noindent where $\w_{\mu,i}$ is the $i$th column of $\W_\mu$. Notably, $\mathbb{E}[z_iz_j] = \mathbb{E}[z_i]\mathbb{E}[z_j] + \text{Cov}(z_i, z_j)$, by the fundamental identity relating the second moment of a random vector to its mean and covariance. Using this identity, we further split the expression into:
\begin{equation}
\begin{split}
    &= \sum_i \sum_j \w_{\mu,i}^T \w_{\mu,j} \mathbb{E}[z_i] \mathbb{E}[z_j] \\
    & \quad
    + \sum_i \sum_j \w_{\mu,i}^T \w_{\mu,j} \text{Cov}(z_i, z_j).
\end{split}
\end{equation}

The first term, containing the expectations, can be factored into the square of a sum: $||\sum_i \w_{\mu,i}\mathbb{E}[z_i]||_2^2$. The second term, involving the covariance is simplified by decomposing the summation into two cases. The first case is when the indices are equal ($i=j$), and the second is when they are not ($i\neq j$):
\begin{equation}
\begin{split}
    & \quad\sum_i \sum_j \w_{\mu,i}^T \w_{\mu,j} \text{Cov}(z_i, z_j) \\
    &= \sum_i ||\w_{\mu,i}||_2^2 \text{Cov}(z_i, z_i) + \sum_{i\neq j} \w_{\mu,i}^T \w_{\mu,j} \text{Cov}(z_i, z_j).
\end{split}
\end{equation}

By definition, the covariance of a variable with itself is its variance: $\text{Cov}(z_i, z_i) = \text{Var}(z_i)$. Additionally, we assume the components of the latent vector $z$ are independent. A standard property of independent random variables is that their covariance is 0. Therefore, for all $i\neq j$, $\text{Cov}(z_i, z_j) = 0$, which cancels out the second summation entirely. By combining the simplified expectation and covariance terms, the final analytical expression for the quadratic term is:
\begin{align}
    \E[||\W_\mu \z||^2_2]
    &= ||\sum_i \w_{\mu,i}\mathbb{E}[z_i]||_2^2  + \sum_i ||\w_{\mu,i}||_2^2 \text{Var}(z_i).
\end{align}

\subsection{Proposition \ref{prop:1}}
\textbf{Proposition 1}. Let $\z\in \mathbb{R}^d$ be a random vector with independent components $z_i$. The first four central moments of each component, $\E[\tilde{z_i}]:=\E[(z_i-\E[z_i])^k]$ for $k\in\{1,2,3,4\}$, can be computed in closed form of the parameters of its distribution if $z_i$ follows:

     1. A Gaussian distribution, $z_i\sim \ensuremath{\mathcal{N}}(\mu_i, \sigma_i^2)$. 
     
     2. A Bernoulli distribution, $z_i\sim \text{Bern}(p_i)$.

\noindent\textbf{Derivation.} 
1. Let $z_i$ be a random variable following a Gaussian distribution, $z_i\sim \mathcal{N}(\mu_i,\sigma_i^2)$. 

The $k$th central moment is defined as $\E[(z_i -\E[z_i])^k]$. \citet{winkelbauer2014momentsabsolutemomentsnormal} introduces the formula to calculate central moments for a Gaussian distribution:
\begin{align}
    \E[(z_i - \mu_i)^k] = 
    \begin{cases} 
    \sigma_i^k (k-1)!! & \text{if } k\in\mathbb{N}_+ \text{ is even,} \\ 
    0 & \text{if } k\in\mathbb{N}_+ \text{ is odd.} 
    \end{cases} 
\end{align}
\noindent where $(k-1)!!$ is the double factorial. Using this formula, we can state the first four central moments:
\begin{itemize}
    \item $k=1$. $\E[z_i - \mu_i] = 0$ since $k$ is odd.
    \item $k=2$. $\E[(z_i - \mu_i)^2] = \sigma_i^2(2-1)!! = \sigma_i^2$ since $k$ is even. Notably, the result is the variance of this Gaussian distribution.
    \item $k=3$. $\E[(z_i - \mu_i)^3] = 0$ since $k$ is odd.
    \item $k=4$. $\E[(z_i - \mu_i)^4] = \sigma_i^4(4-1)!! = 3\sigma_i^4$ since $k$ is even.
\end{itemize}

Therefore, all four central moments are closed-form functions of the distribution's variance $\sigma_i^2$.

2. Let $z_i$ be a random variable following a Bernoulli distribution, $z_i\sim \text{Bern}(p_i)$. The variable $z_i$ takes the value 1 with probability $p_i$ and 0 with probability $1-p_i$. The mean is $\E[z_i] =p_i$.

The $k$th central moment is defined as $\E[(z_i -\E[z_i])]$. We consider the two possible outcomes for $z_i$:
\begin{itemize}
    \item If $z_i=1$, then $(z_i-p_i)^k = (1-p_i)^k$.
    \item If $z_i=0$, then $(z_i-p_i)^k = (-p_i)^k$.
\end{itemize}

We can compute the central moments using the definition of expectation.
\begin{align}
    \E[(z_i - p_i)^k] = (1-p_i)^k \cdot p_i + (-p_i)^k \cdot (1-p_i).
\end{align}

Now we can compute the first four central moments:
\begin{itemize}
    \item $k=1$. $\E[z_i - p_i] = (1-p_i)^1 p_i + (-p_i)^1 (1-p_i) = p_i - p_i^2 - p_i + p_i^2 = 0$.
    \item $k=2$.  
    \begin{align}
        \E[(z_i - p_i)^2] &= (1-p_i)^2 p_i + (-p_i)^2 (1-p_i), \notag\\
        &= p_i-2p_i^2+p_i^3+p_i^2-p_i^3, \notag\\
        &= p_i-p_i^2 = p_i(1-p_i).
    \end{align}
    \item $k=3$. 
    \begin{align}
        \E[(z_i - p_i)^3] &= (1-p_i)^3 p_i + (-p_i)^3 (1-p_i), \notag\\
        &= (1-3p_i+3p_i^2-p_i^3)p_i - p_i^3(1-p_i),  \notag\\
        &= p_i - 3p_i^2 + 3p_i^3 - p_i^4 - p_i^3 + p_i^4, \notag\\
        &= p_i - 3p_i^2 + 2p_i^3 \notag\\
        &= p_i(1-p_i)(1-2p_i).
    \end{align}
    \item $k=4$. 
    \begin{align}
        \E[(z_i - p_i)^4] &= (1-p_i)^4 p_i + (-p_i)^4 (1-p_i), \notag\\
        &= (1-4p_i+6p_i^2-4p_i^3+p_i^4)p_i \notag\\
        & \quad + p_i^4(1-p_i),  \notag\\
        &= p_i - 4p_i^2 + 6p_i^3 - 4p_i^4 + p_i^5 + p_i^4 - p_i^5, \notag\\
        &= p_i - 4p_i^2 + 6p_i^3 - 3p_i^4 \notag\\
        &= p_i(1-p_i)(1-3p_i+3p_i^2).
    \end{align}
\end{itemize}

Therefore, all four central moments are closed-form functions of the parameter $\p_i$.

\subsection{Theorem \ref{thm:1}}
\textbf{Theorem 1.} Let $\W_\mu \z$ and $\W_\alpha \z$ be two linear projections of a random vector $\z$ whose components $z_i$ are independent. The covariance terms $\text{Cov}(\W_\mu \z, ||\W_\alpha \z||^2)$ and $\text{Cov}(||\W_\mu \z||_2^2, ||\W_\alpha \z||_2^2)$ can be expressed as a linear combination of the first four central moments of the components $z_i$, respectively. The coefficients of this linear combination are polynomials in the entries of the matrices $\W_\mu$ and $\W_\alpha$.

\noindent\textbf{Proof.} For simplicity in writing, we define:
\begin{align}
    & u_1 = u_2 =  \W_\mu \z, \quad v_1 = v_2 =\W_\alpha \z, \\
    & \Delta u_1 = \Delta u_2 = \W_\mu \z - \mathbb{E}[\W_\mu \z] = \W_\mu (\z-\mathbb{E}[\z]), \\
    & \Delta v_1 = \Delta v_2 = \W_\alpha \z - \mathbb{E}[\W_\alpha \z] = \W_\alpha  (\z-\mathbb{E}[\z]).
\end{align}

And we denote $\tilde{\z} =\z -\mathbb{E}[\z]$, thus $\mathbb{E}[\tilde{\z}] = 0$, and $\E[(\tilde{\z})^2] = \text{Var}(\z)$.

\citet{covariance} introduces the formula to compute covariance between the products of independent variables as follows:

\begin{align}
\begin{split}
    \text{Cov}(u_1, v_1v_2) &= \mathbb{E}[v_1]\text{Cov}(v_2,u_1) + \mathbb{E}[v_2]\text{Cov}(v_1,u_1) \\
    &\quad + \mathbb{E}[(\Delta v_1)(\Delta v_2)(\Delta u_1)].
\end{split}
\end{align}

Identical to the fixed variance case, we can derive $\text{Cov}(v_1,u_1) = \text{Cov}(v_2,u_1) = \sum_i \w_{\alpha, i}^T \w_{\mu, i}\text{Var}(z_i) = \W_\alpha^T \W_\mu \E[(\tilde{\z})^2]$. And to compute the last term, we expand it as follows, with $\odot$ denoting Hadamard product:

\begin{align}
        & \quad\mathbb{E}\bigl[(\Delta v_1)(\Delta v_2)(\Delta u_1)\bigr] \\
        &= \mathbb{E}\bigl[(\W_\mu \tilde{\z})\odot(\W_\alpha\tilde{\z})\odot(\W_\alpha\tilde{\z})\bigr], \\
        &= \mathbb{E}\bigl[\sum_i\sum_j\sum_k(\w_{\mu, i}\tilde{z}_i)\odot(\w_{\alpha, j}\odot\w_{\alpha, k}\tilde{z}_j\tilde{z}_k)\bigr],\\
        & = \sum_i\sum_j\sum_k \w_{\mu, i}\odot\w_{\alpha, j}\odot\w_{\alpha, k}\mathbb{E}[\tilde{z}_i\tilde{z}_j\tilde{z}_k].
    \end{align}

Notably, $\mathbb{E}[\tilde{z}_i\tilde{z}_j\tilde{z}_k]$ is nonzero only if $i=k=j$. In other generic cases, for example, $i=j\neq k$, we can always separate $\tilde{z}_i\tilde{z}_j$ from $\tilde{z}_k$. In fact, because of the constraint, we can simplify the expression:
    \begin{align}
        \mathbb{E}[\tilde{z}_i\tilde{z}_i\tilde{z}_k] &= \mathbb{E}[(\tilde{z}_i)^2\tilde{z}_k] \\
        &=\mathbb{E}[(\tilde{z}_i)^2]\mathbb{E}[\tilde{z}_k] \\
        &=0
    \end{align}

    Therefore, the only case we need to consider is when $i=j=k$, and thus we can write:
    \begin{align}
        \mathbb{E}\bigl[(\Delta v_1)(\Delta v_2)(\Delta u_1)\bigr] &=  \sum_i\sum_j\sum_k \w_{\mu, i}\odot\w_{\alpha, j}\odot\w_{\alpha, k}\notag\\
        &\quad \mathbb{E}[\tilde{z}_i\tilde{z}_i\tilde{z}_i] \\
        &= \sum_i\w_{\mu, i}\odot\w_{\alpha, j}\odot\w_{\alpha, k}\mathbb{E}[(\tilde{z_i})^3]
    \end{align}

Piecing all together, we derive the expression for the covariance term $\text{Cov}(\W_\mu \z, (\W_\alpha \z)^2)$:

\begin{equation}
\label{eq:cov_1_2}
\begin{split}
        \text{Cov}(\W_\mu \z, ||\W_\alpha \z||_2^2) &= (\W_\alpha\mathbb{E}[\z])\odot (\W_\alpha\odot\W_\mu \mathbb{E}[(\tilde{\z})^2]) \\
        & \quad+ (\W_\alpha\mathbb{E}[\z])\odot( \W_\alpha \odot\W_\mu \mathbb{E}[(\tilde{\z})^2]) \\
        & \quad \sum_i\w_{\mu, i}\odot\w_{\alpha, j}\odot\w_{\alpha, k}\mathbb{E}[(\tilde{z_i})^3], \\
        &=2(\W_\alpha\mathbb{E}[\z])\odot(\W_\alpha \odot\W_\mu \mathbb{E}[(\tilde{\z})^2])  \\
        & \quad+ \sum_i\w_{\mu, i}\odot\w_{\alpha, j}\odot\w_{\alpha, k}\mathbb{E}[(\tilde{z_i})^3].
\end{split}
\end{equation}

\citet{covariance} also introduced the formula to calculate the covariance between two products of random variables:

\begin{align}
\label{eq:cov_2_2}
    &\quad \Cov(u_1u_2,v_1v_2) 
    \notag\\
    &= \mathbb{E}(u_1)\,\mathbb{E}(v_1)\,\Cov(u_2,v_2)
               + \mathbb{E}(u_1)\,\mathbb{E}(v_2)\,\Cov(u_2,v_1)
               \notag\\
                &\quad + \mathbb{E}(u_2)\,\mathbb{E}(v_1)\,\Cov(u_1,v_2)
               + \mathbb{E}(u_2)\,\mathbb{E}(v_2)\,\Cov(u_1,v_1)
    \notag\\
    &\quad + \mathbb{E}\bigl[\Delta u_1\,\Delta u_2\,\Delta v_1\,\Delta v_2\bigr]
           + \mathbb{E}(u_1)\,\mathbb{E}\bigl[\Delta u_2\,\Delta v_1\,\Delta v_2\bigr]
     \notag\\
    &\quad        
           + \mathbb{E}(u_2)\,\mathbb{E}\bigl[\Delta u_1\,\Delta v_1\,\Delta v_2\bigr] + \mathbb{E}(v_1)\,\mathbb{E}\bigl[\Delta u_1\,\Delta u_2\,\Delta v_2\bigr]
    \notag\\
    &\quad
           + \mathbb{E}(v_2)\,\mathbb{E}\bigl[\Delta u_1\,\Delta u_2\,\Delta v_1\bigr]
           - \Cov(u_1,u_2)\,\Cov(v_1,v_2).
    \end{align}

Following the derivation earlier, we can compute the terms $\mathbb{E}(u_1)\,\mathbb{E}\bigl[\Delta u_2\,\Delta v_1\,\Delta v_2\bigr]$, $\mathbb{E}(u_2)\,\mathbb{E}\bigl[\Delta u_1\,\Delta v_1\,\Delta v_2\bigr]$, $\mathbb{E}(v_1)\,\mathbb{E}\bigl[\Delta u_1\,\Delta u_2\,\Delta v_2\bigr]$, $\mathbb{E}\bigl[\Delta u_1\,\Delta u_2\,\Delta v_1\bigr].$ 
And similarly, we wish to consider all nonzero cases in $ \mathbb{E}\bigl[\Delta u_1\,\Delta u_2\,\Delta v_1\,\Delta v_2\bigr]$, and they are: $i=j=k=l, i=j\neq k=l, i=k\neq j=l, i=l\neq k=j$.
    \begin{align}
        &\quad \mathbb{E}\bigl[\Delta u_1\,\Delta u_2\,\Delta v_1\,\Delta v_2\bigr] \\
        &= \mathbb{E}\bigl[\W_\mu \tilde{z} \odot\W_\mu \tilde{z}\odot\W_\alpha\tilde{z}\odot\W_\alpha\tilde{z}\bigr] \\
        &= \mathbb{E}\bigl[\sum_i\sum_j\sum_k\sum_l\left[(\w_{\mu, i}\tilde{z}_i)\odot(\w_{\mu, j}\tilde{z}_j)\right]\odot \notag \\
        & \quad\left[(\w_{\alpha, k}\tilde{z}_k)\odot(\w_{\alpha, l}\tilde{z}_l)\right]\bigr]\\
        & = \sum_i\sum_j\sum_k \sum_j \w_{\mu, i} \odot \w_{\mu, j}\odot\w_{\alpha, k}\odot\w_{\alpha, l}\mathbb{E}[\tilde{z}_i\tilde{z}_j\tilde{z}_k\tilde{z}_l]
    \end{align}

Consider the case where $i=j=k=l$,
    \begin{align}
        & \quad \w_{\mu, i} \odot \w_{\mu, i}\odot\w_{\alpha, i}\odot\w_{\alpha, i}\mathbb{E}[\tilde{z}_i\tilde{z}_i\tilde{z}_i\tilde{z}_i] \notag \\
        &=\sum_i \w_{\mu, i} \odot \w_{\mu, i}\odot\w_{\alpha, i}\odot\w_{\alpha, i}\mathbb{E}[(\tilde{z_i})^4]
    \end{align}

    And consider the case where $i=j\neq k=l$
    \begin{align}
         &\quad \sum_i\sum_{j, j\neq i}  (\w_{\mu, i} \odot \w_{\mu, i})\odot(\w_{\alpha, j}\odot\w_{\alpha, j})\mathbb{E}[\tilde{z}_i\tilde{z}_i\tilde{z}_j\tilde{z}_j] \\
         &= \sum_i\sum_{j, j\neq i} (\w_{\mu, i} \odot \w_{\mu, i})\odot(\w_{\alpha, j}\odot\w_{\alpha, j})\mathbb{E}[(\tilde{z}_i)^2 (\tilde{z}_j)^2] \\
         &= \sum_i\sum_{j, j\neq i} (\w_{\mu, i} \odot \w_{\mu, i})\odot(\w_{\alpha, j}\odot\w_{\alpha, j})\mathbb{E}[(\tilde{z}_i)^2]\mathbb{E}[(\tilde{z}_j)^2] \\
         &= \sum_i (\w_{\mu, i} \odot \w_{\mu, i})\mathbb{E}[(\tilde{z}_i)^2]\sum_{j, j\neq i}(\w_{\alpha, j} \odot \w_{\alpha, j})\mathbb{E}[(\tilde{z}_j)^2] \\
         &= \sum_i (\w_{\mu, i} \odot \w_{\mu, i})\mathbb{E}[(\tilde{z}_i)^2]\notag\\
         & \quad \left(\sum_{j} (\w_{\alpha, j} \odot \w_{\alpha, j})\mathbb{E}[(\tilde{z}_j)^2] - (\w_{\alpha, i} \odot \w_{\alpha, i})\mathbb{E}[(\tilde{z}_i)^2] \right)\\
         &=\sum_i (\w_{\mu, i} \odot \w_{\mu, i})\mathbb{E}[(\tilde{z}_i)^2]\sum_{j}(\w_{\alpha, j} \odot \w_{\alpha, j})\mathbb{E}[(\tilde{z}_j)^2] \notag\\
         &\quad - \sum_i (\w_{\mu, i} \odot \w_{\mu, i})\mathbb{E}[(\tilde{z}_i)^2](\w_{\alpha, i} \odot \w_{\alpha, i})\mathbb{E}[(\tilde{z}_i)^2]
    \end{align}

    This holds because we know $i\neq k$. Similarly, when $i=k\neq j=l,$

    \begin{align}
         &\quad \sum_i\sum_{j,j\neq i}  (\w_{\mu, i} \odot \w_{\mu, j})\odot(\w_{\alpha, i} \odot \w_{\alpha, j})\mathbb{E}[\tilde{z}_i\tilde{z}_j\tilde{z}_i\tilde{z}_j] \\
         &= \sum_i\sum_{j,j\neq i} (\w_{\mu, i} \odot \w_{\mu, j})\odot(\w_{\alpha, i} \odot \w_{\alpha, j})\mathbb{E}[(\tilde{z}_i)^2 (\tilde{z}_j)^2] \\
         &= \sum_i\sum_{j,j\neq i} (\w_{\mu, i} \odot \w_{\mu, j})\odot(\w_{\alpha, i} \odot \w_{\alpha, j})\mathbb{E}[(\tilde{z}_i)^2]\mathbb{E}[ (\tilde{z}_j)^2] \\
         &= \sum_i \w_{\mu, i}\odot\w_{\alpha, i}\mathbb{E}[(\tilde{z}_i)^2]\sum_{j,j\neq i} \w_{\mu, j}\odot\w_{\alpha, j}\mathbb{E}[ (\tilde{z}_j)^2]\\
         &= \sum_i \w_{\mu, i}\odot\w_{\alpha, i}\mathbb{E}[(\tilde{z}_i)^2]\sum_j \w_{\mu, j}\odot\w_{\alpha, j}\mathbb{E}[(\tilde{z}_j)^2] \notag\\
         & \quad - \sum_i\w_{\mu, i}\odot\w_{\alpha, i}\mathbb{E}[ (\tilde{z}_i)^2] \w_{\mu, i}\odot\w_{\alpha, i}\mathbb{E}[ (\tilde{z}_i)^2]
    \end{align}

    When $i=l\neq k=j$,

    \begin{align}
         &\quad \sum_i\sum_{j,j\neq i}  (\w_{\mu, i}\odot \w_{\mu, j})\odot(\w_{\alpha, j}\odot\w_{\alpha, i})\mathbb{E}[\tilde{z}_i\tilde{z}_j\tilde{z}_j\tilde{z}_i] \\
         &= \sum_i\sum_j (\w_{\mu, i}\odot \w_{\mu, j})\odot(\w_{\alpha, j}\odot\w_{\alpha, i})\mathbb{E}[(\tilde{z}_i)^2 (\tilde{z}_j)^2] \\
         &= \sum_i\sum_{j,j\neq i} (\w_{\mu, i}\odot \w_{\mu, j})\odot(\w_{\alpha, j}\odot\w_{\alpha, i})\mathbb{E}[(\tilde{z}_i)^2]\mathbb{E}[ (\tilde{z}_j)^2] \\
         &= \sum_i \w_{\mu, i}\odot\w_{\alpha, i}\mathbb{E}[(\tilde{z}_i)^2]\sum_{j,j\neq i} \w_{\mu, j}\odot\w_{\alpha, j}\mathbb{E}[ (\tilde{z}_j)^2]\\
         &= \sum_i \w_{\mu, i}\odot\w_{\alpha, i}\mathbb{E}[(\tilde{z}_i)^2]\notag\\
         & \quad \left(\sum_{j} \w_{\mu, j}^T\w_{\alpha, j}\mathbb{E}[ (\tilde{z}_j)^2] - \w_{\mu, i}\odot\w_{\alpha, i}\mathbb{E}[ (\tilde{z}_i)^2] \right) \\
         &= \sum_i \w_{\mu, i}\odot\w_{\alpha, i}\mathbb{E}[(\tilde{z}_i)^2]\sum_{j} \w_{\mu, j}\odot\w_{\alpha, j}\mathbb{E}[ (\tilde{z}_j)^2] \notag\\
         &\quad
        - \sum_i\w_{\mu, i}\odot\w_{\alpha, i}\mathbb{E}[ (\tilde{z}_i)^2] \w_{\mu, i}\odot\w_{\alpha, i}\mathbb{E}[ (\tilde{z}_i)^2] \\
    \end{align}

    Since these four cases are mutually exclusive, we could rewrite the full term as: 

    \begin{align}
        &\quad \mathbb{E}\bigl[\Delta u_1\,\Delta u_2\,\Delta v_1\,\Delta v_2\bigr] \notag\\
        &= \sum_i \w_{\mu, i} \odot \w_{\mu, i}\odot\w_{\alpha, i}\odot\w_{\alpha, i}\mathbb{E}[(\tilde{z_i})^4] \notag\\
        &\quad + \sum_i \w_{\mu,i}\odot\w_{\mu,i}\mathbb{E}[(\tilde{z}_i)^2]\sum_{j}\w_{\alpha,j}\odot\w_{\alpha,j}\mathbb{E}[(\tilde{z}_j)^2]\notag\\
        &\quad - \sum_i\w_{\mu, i}\odot\w_{\alpha, i}\mathbb{E}[ (\tilde{z}_i)^2]\sum_j\w_{\mu, j}\odot\w_{\alpha, j}\mathbb{E}[ (\tilde{z}_j)^2] \notag\\
        &\quad + 2\sum_i \w_{\mu, i}\odot\w_{\alpha, i}\mathbb{E}[(\tilde{z}_i)^2]\sum_j \w_{\mu, j}\odot\w_{\alpha, j}\mathbb{E}[(\tilde{z}_j)^2]
        \notag\\
    &\quad
     - 2\sum_i(\w_{\mu, i}\odot\w_{\alpha, i}\mathbb{E}[ (\tilde{z}_i)^2])(\w_{\mu, i}\odot\w_{\alpha, i}\mathbb{E}[ (\tilde{z}_i)^2]), \\
    &= \sum_i \w_{\mu, i} \odot \w_{\mu, i}\odot\w_{\alpha, i}\odot\w_{\alpha, i}\mathbb{E}[(\tilde{z_i})^4]  \notag\\  
    &\quad + \sum_i \w_{\mu, i} \odot \w_{\mu, i}\mathbb{E}[(\tilde{z}_i)^2]\sum_{j}\w_{\alpha, i}\odot\w_{\alpha, i}\mathbb{E}[(\tilde{z}_j)^2]
        \notag\\
    &\quad
     + 2\sum_i \w_{\mu, i}\odot\w_{\alpha, i}\mathbb{E}[(\tilde{z}_i)^2]\sum_j \w_{\mu, j}\odot\w_{\alpha, j}\mathbb{E}[(\tilde{z}_j)^2]\notag\\
     &\quad - 3\sum_i(\w_{\mu, i}\odot\w_{\alpha, i}\mathbb{E}[ (\tilde{z}_i)^2])(\w_{\mu, i}\odot\w_{\alpha, i}\mathbb{E}[ (\tilde{z}_i)^2]).
    \end{align}

Putting this back to \Cref{eq:cov_2_2}, we can write the final expression as:

\begin{align}
\label{eq:covariance_2}
    &\quad \Cov(u_1u_2,v_1v_2) = \text{Cov}(||\W_\mu z||_2^2, ||\W_\alpha z||_2^2)\notag\\
    &= 4(\W_\mu\odot\W_\mu\odot\W_\alpha\odot\W_\alpha(\mathbb{E}[\z])^2\mathbb{E}[(\tilde{\z})^2]  \notag\\
    &\quad +\sum_i \w_{\mu, i} \odot \w_{\mu, i}\odot\w_{\alpha, i}\odot\w_{\alpha, i}\mathbb{E}[(\tilde{z_i})^4] 
    \notag\\
    &\quad
    + 2\sum_i \w_{\mu, i}\odot\w_{\alpha, i}\mathbb{E}[(\tilde{z}_i)^2]\sum_j \w_{\mu, j}\odot\w_{\alpha, j}\mathbb{E}[(\tilde{z}_j)^2]\notag\\
     &\quad - 3\sum_i(\w_{\mu, i}\odot\w_{\alpha, i}\mathbb{E}[ (\tilde{z}_i)^2])(\w_{\mu, i}\odot\w_{\alpha, i}\mathbb{E}[ (\tilde{z}_i)^2]) \notag\\
    &\quad
           + 2\W_\mu\mathbb{E}[z]\odot \sum_i \w_{\mu, i}\odot\w_{\alpha, i}\odot\w_{\alpha, i}\mathbb{E}[(\tilde{z_i})^3]
    \notag\\
    &\quad + 2\W_\alpha\mathbb{E}[z]\odot \sum_i \w_{\mu, i}\odot \w_{\mu, i}\odot\w_{\alpha, i}\mathbb{E}[(\tilde{z_i})^3].
    \end{align}

Therefore, we show that the covariance terms $\text{Cov}(\W_\mu \z, ||\W_\alpha \z||^2)$ (cf. Eq.~\ref{eq:cov_1_2}) and $\text{Cov}(||\W_\mu \z||_2^2, ||\W_\alpha \z||_2^2)$ (cf. Eq.~\ref{eq:covariance_2}) can be expressed as a linear combination of the first four central moments of the components $z_i$. The coefficients of this linear combination are polynomials in the entries of the weight matrices $\W_\mu$ and $\W_\alpha$.

\section{Experiment Details}
\label{appendix:experiment_details}
\subsection{Uniform Dequantization}
Our model's decoder defines a likelihood over continuous values using a Gaussian distribution. However, the image datasets we use, such as MNIST, consist of discrete pixels values. To bridge this gap, we employ uniform dequantization. This standard technique adds a small amount of uniform noise to each discrete pixel value, transforming data into a continuous variable that is compatible with our model's likelihood function. 

Specifically, for discrete data $\x_\text{int}$ with values in $\{0,1,\dots, 255\}$, the dequantized data is defined as $\y=\x+\varu$, where $\x = \frac{\x_\text{int}}{256}$, $\varu\sim \ensuremath{\mathcal{U}}[0,\frac{1}{256})$. This process maps each discrete pixel value to a unique continuous bin of width $\frac{1}{256}$ within $[0,1)$.

The true probability of a discrete pixel value under our continuous model is thus defined as:

\begin{align}
    P_\text{model}(X = \x_{\text{int}}) = \int^{(\x_{\text{int}}+1)/256}_{\x_{\text{int}}/256} p_{\text{model}}(\y) d\y
\end{align}

By applying Jensen's inequality, we can establish a formal relationship:
\begin{align}
    \E_{\varu}[\log p_{\text{model}}(\y)] &\le \log \left( \mathbb{E}_{\varu}[p_{\text{model}}(\y)] \right), \\
    &= \log P_{\text{model}}(X = \x_{\text{int}}) - \log(256).
\end{align}

Rearranging this gives us a lower bound on the discrete log-likelihood:

\begin{align}
    \log P_{\text{model}}(X = \x_{\text{int}}) \ge \E_{\varu}[\log p_{\text{model}}(\y)] + \log(256).
\end{align}

Therefore, to ensure we are optimizing a valid lower bound on the true log-likelihood of the discrete data, we apply a correction to the pixel-wise reconstruction log-likelihood by adding a constant $\log(256)$ to it.

\subsection{Model Architecture}

In this section, we detail the model architecture used in the experiments in section 3 and 4.

\subsubsection{Fixed Variance Experiment}

In section 3, we conduct a controlled experiment with a fixed output variance, the VAE consists of a convolutional encoder and a simple linear decoder. The encoder architecture, which is shared across both continuous and discrete latent space models, is detailed in \Cref{tab:encoder_arch}. The decoder is a single fully-connected linear layer that maps the latent variable $\z$ directly to the flattened output image pixels. Equivalent to a learnable bias, we augment the latent vector $\z$ by concatenating it with an additional dimension fixed at a constant value of $1$.

\begin{table}[h]
\centering
\caption{Encoder architecture for the fixed and learnable variance experiments on MNIST.}
\label{tab:encoder_arch}
\begin{tabular}{lcccc}
\toprule
\textbf{Layer} & \textbf{Kernel Size} & \textbf{Stride} & \textbf{Padding} & \textbf{Activation} \\
\midrule
Conv2d & 3x3 & 1 & 1 & ReLU \\
Conv2d & 3x3 & 1 & 1 & ReLU \\
Conv2d & 3x3 & 1 & 1 & - \\
Flatten & - & - & - & - \\
Linear & - & - & - & - \\
\bottomrule
\end{tabular}
\end{table}

\subsubsection{Learnable Variance Experiment}

\paragraph{MNIST. } The encoder for the MNIST experiments is consistent with fixed-variance experiment, as presented in \Cref{tab:encoder_arch}. The nonlinear decoder mirrors this structure with a linear layer followed by several convolutional layers. The architecture is detailed in \Cref{tab:mnist_learnable_decoder_arch}. The linear decoder is a single fully-connected layer without any activations.

\begin{table}[h]
\centering
\caption{Nonlinear Decoder architecture for the learnable variance experiments on MNIST.}
\label{tab:mnist_learnable_decoder_arch}
\begin{tabular}{lcccc}
\toprule
\textbf{Layer} & \textbf{Kernel Size} & \textbf{Stride} & \textbf{Padding} & \textbf{Activation} \\
\midrule
Linear & - & - & - & - \\
Reshape & - & - & - & - \\
Conv2d & 3x3 & 1 & 1 & ReLU \\
Conv2d & 3x3 & 1 & 1 & ReLU \\
Conv2d & 3x3 & 1 & 1 & ReLU \\
Conv2d & 3x3 & 1 & 1 & ReLU \\
Conv2d & 1x1 & 1 & 0 & - \\
\bottomrule
\end{tabular}
\end{table}

\paragraph{ImageNet Architecture and CIFAR-10}

For the more complex CIFAR-10 and ImageNet datasets, we use deeper, strided convolutional architectures for both encoder and the nonlinear decoder, with batch normalization after each convolutional layers. The linear decoder remains a single fully-connected layers. The architectures are detailed in \Cref{tab:cifar_encoder_arch} and \Cref{tab:cifar_decoder_arch}.

\begin{table}[h]
\centering
\small
\caption{Encoder architecture for the learnable variance experiments on CIFAR-10 and ImageNet.}
\label{tab:cifar_encoder_arch}
\begin{tabular}{lcccc}
\toprule
\textbf{Layer} & \textbf{Kernel Size} & \textbf{Stride} & \textbf{Padding} & \textbf{Activation} \\
\midrule
Conv2d & 4x4 & 2 & 1 & ReLU \\
BatchNorm2d & - & - & - & - \\
Conv2d & 4x4 & 2 & 1 & ReLU \\
BatchNorm2d & - & - & - & - \\
Conv2d & 4x4 & 2 & 1 & ReLU \\
BatchNorm2d & - & - & - & - \\
Conv2d & 4x4 & 1 & 0 & ReLU \\
BatchNorm2d & - & - & - & - \\
Flatten & - & - & - & - \\
Linear & - & - & - & - \\
\bottomrule
\end{tabular}
\end{table}

\begin{table}[h]
\centering
\small
\caption{Nonlinear Decoder architecture for the learnable variance experiments on CIFAR-10 and ImageNet.}
\label{tab:cifar_decoder_arch}
\begin{tabular}{lcccc}
\toprule
\textbf{Layer} & \textbf{Kernel Size} & \textbf{Stride} & \textbf{Padding} & \textbf{Activation} \\
\midrule
Linear & - & - & - & - \\
Reshape & - & - & - & - \\
ConvTranspose2d & 4x4 & 1 & 0 & ReLU \\
BatchNorm2d & - & - & - & - \\
ConvTranspose2d & 4x4 & 2 & 1 & ReLU \\
BatchNorm2d & - & - & - & - \\
ConvTranspose2d & 4x4 & 2 & 1 & ReLU \\
BatchNorm2d & - & - & - & - \\
ConvTranspose2d & 4x4 & 2 & 1 & - \\
\bottomrule
\end{tabular}
\end{table}

\subsection{Training Details}

\subsubsection{Data Preprocessing}
For all experiments, the input images are first transformed into PyTorch tensors. Before being passed to the model, the image data is scaled by $\frac{255}{256}$, in preparation for uniform dequantization.

\subsubsection{Baselines}
For our baseline models used throughout the following experiments, we use standard implementations for the Gumbel-Softmax and the reparameterization trick in VAEs. For the REINFORCE, we implement a baseline to reduce variance. Specifically, we use the running average of the reconstruction loss.

\subsubsection{Fixed Variance Experiment}
All models in this experiment are trained using the AdamW optimizer with betas set to $(0.9, 0.95)$. We performed a hyperparameter search for the learning rate over the values $\{1\times 10^{-4}, 5\times 10^{-4}, 1\times 10^{-5}, 5\times 10^{-5}\}$. For each model, the best performing rate was selected based on the final BPD score, which was evaluated on the validation set at the end of Epoch 500.  All reported metrics in \Cref{tab:mnist_results} are likewise evaluated on the validation set at this same epoch.

The output variance of the decoder was fixed for these experimented. We tested $\sigma^2$ values of $\{0.1, 0.05, 0.01\}$ and found that a fixed variance $\sigma^2 = 0.01$ yielded the best results across all models. The models are trained with a batch size of 64, and no gradient clipping was applied. Additionally, we did not use KL annealing; the $\beta$ parameter for KLD is fixed at 1.0 throughout training.

\paragraph{Gradient Variance Calculation} The gradient variance with respect to the encoder parameters  reported in \Cref{tab:log_variance} is measured empirically. To isolate the variance only from the latent variable sampling, we first perform a single forward pass through the encoder on a fixed batch of data to obtain the parameters of the latent distribution, which is to avoid the randomness introduced by the uniform noise we add for dequantization. With these parameters held constant, we then draw 100 latent samples from this fixed distribution. For each sample, we compute the corresponding reconstruction loss and backpropagate to get a gradient vector with respect to the encoder's parameters. The total gradient variance is computed by first calculating the variance for each individual parameter in the encoder across the 100 gradient samples. These per-parameter variances are then summed together to produce the final scalar value that is reported in \Cref{tab:log_variance}.

\subsubsection{Learnable Variance Experiment}

The training scheme is similar to the fixed variance experiment. All models are trained using AdamW optimizer(s) with betas of $(0.9, 0.95)$, and we selected the best learning rate for each baseline from the set $\{1\times 10^{-4}, 5\times 10^{-4}, 1\times 10^{-5}, 5\times 10^{-5}\}$.
For our combined methods that integrate Silent Gradients, we built upon the best baseline learning rates and introduced separate optimizers for the linear decoder's $\mu$ and $\alpha$ components. We perform a hyperparameter search for the annealing rate from $\{1\times10^{-2}, 5\times10^{-2}, 1\times10^{-3}, 5\times10^{-3}\}$, and the encoder freeze epoch (cut-off) from $\{50, 80, 100, 150, 200\}$.

The training duration and batch sizes varied by dataset:
\paragraph{MNIST. } Models are trained with a batch size of 64. The REINFORCE models are trained for 300 epochs, while all others are trained for 200 epochs.
\paragraph{ImageNet. } Models are trained for 100 epochs with a batch size of 128.
\paragraph{CIFAR-10. } Models are trained for 2000 epochs with a batch size of 256.

At the end of training, all models are guaranteed to be converged. And like the fixed variance experiment, no KL annealing or $\beta$ parameter for KLD other than 1.0 is used.

\subsubsection{Computing Resources}
All experiments included were conducted on 8 NVIDIA GeForce RTX 4090 GPUs.

\subsubsection{Additional Results}

As a complement to \Cref{tab:learnable_var_results}, we report the corresponding MSE results from a representative single run to provide an additional view of reconstruction quality, as shown in \Cref{tab:learnable_var_mse}.

\begin{table*}[ht]
\centering
\caption{Performance comparison in MSE ($\downarrow$) for models with learnable variance across datasets. Results are reported from a single run. For each baseline, the lower MSE between ``Without SG'' and ``With SG'' is shown in bold.}
\label{tab:learnable_var_mse}
\resizebox{\linewidth}{!}{%
\begin{tabular}{llcccccc}
\toprule
& \multirow{2}{*}[-0.3em]{Method} 
& \multicolumn{2}{c}{MNIST} 
& \multicolumn{2}{c}{ImageNet} 
& \multicolumn{2}{c}{CIFAR-10} \\
\cmidrule(lr){3-4} \cmidrule(lr){5-6} \cmidrule(lr){7-8}
& & Without SG & With SG 
& Without SG & With SG 
& Without SG & With SG \\
\midrule
\multirow{2}{*}{Continuous}
& Silent Gradients (SG)
& -- 
& 1.79 
& -- 
& 17.30 
& -- 
& 10.28 \\

& Reparameterization 
& 1.36 
& \textbf{1.60} 
& 14.56 
& \textbf{12.69} 
& 12.54 
& \textbf{8.18} \\
\midrule
\multirow{3}{*}{Discrete}
& Silent Gradients (SG)
& -- 
& 10.66 
& -- 
& 28.15 
& -- 
& 39.56 \\

& Gumbel-Softmax 
& 5.63 
& \textbf{6.02} 
& 29.34 
& \textbf{23.04} 
& 21.12 
& \textbf{20.36} \\

& REINFORCE 
& 11.59 
& \textbf{8.20} 
& 50.56 
& \textbf{43.25} 
& 41.50 
& \textbf{36.26} \\
\bottomrule
\end{tabular}
}
\end{table*}

\section{Visualized Output}
\label{appendix:visualized_output}
In this section, we provide the visualization of reconstructed images using the trained model at the epoch where the metrics are reported.

The visualizations are generated by taking a fixed batch of images from the validation set of each respective dataset. These images are passed through the train encoder to obtain a latent variable $\z$, which is then passed to the corresponding decoder to produce the reconstruction. For the learnable variable experiment in which a dual-decoder setting is introduced, only nonlinear decoder is used to generate the reconstruction. The linear decoder used for computing the Silent Gradient, in this case, is not used.

\subsection{Fixed Variance Experiment}
\begin{figure}[ht]
    \centering
        \includegraphics[width=0.48\textwidth]{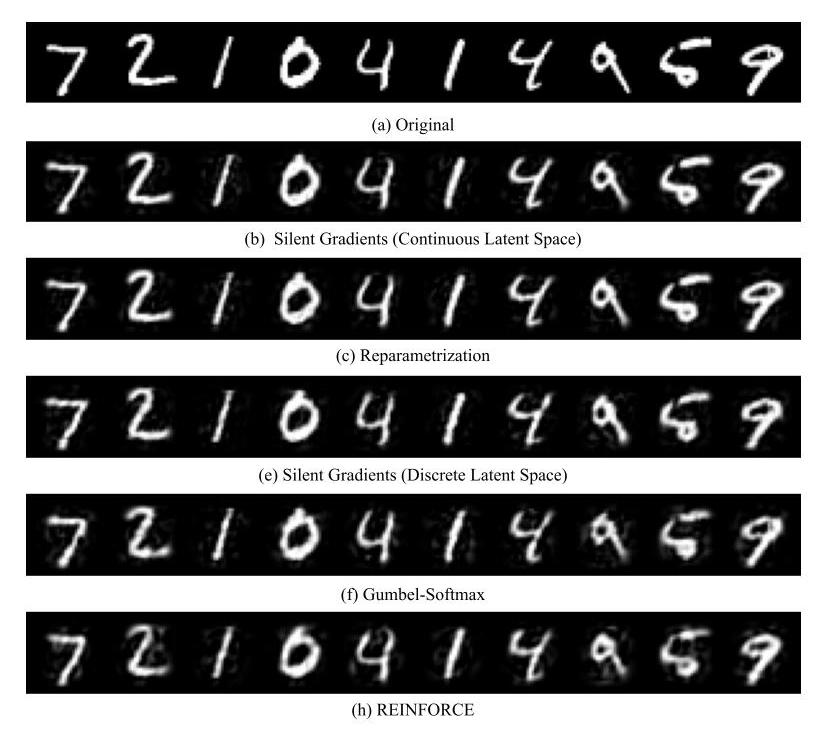}
        \caption{Visual comparison of reconstructions for the fixed variance experiment on MNIST. The top row (a) displays original images from the validation set. Subsequent rows show the reconstructed means from our Silent Gradients method and the baseline estimators for both continuous and discrete latent spaces.}
        \label{fig:mnist_fixed}
\end{figure}

The visualizations show the original images and the corresponding reconstructed mean, as in \Cref{fig:mnist_fixed}.

\subsection{Learnable Variance Experiment}
\begin{figure}[ht]
    \centering
        \includegraphics[width=0.48\textwidth]{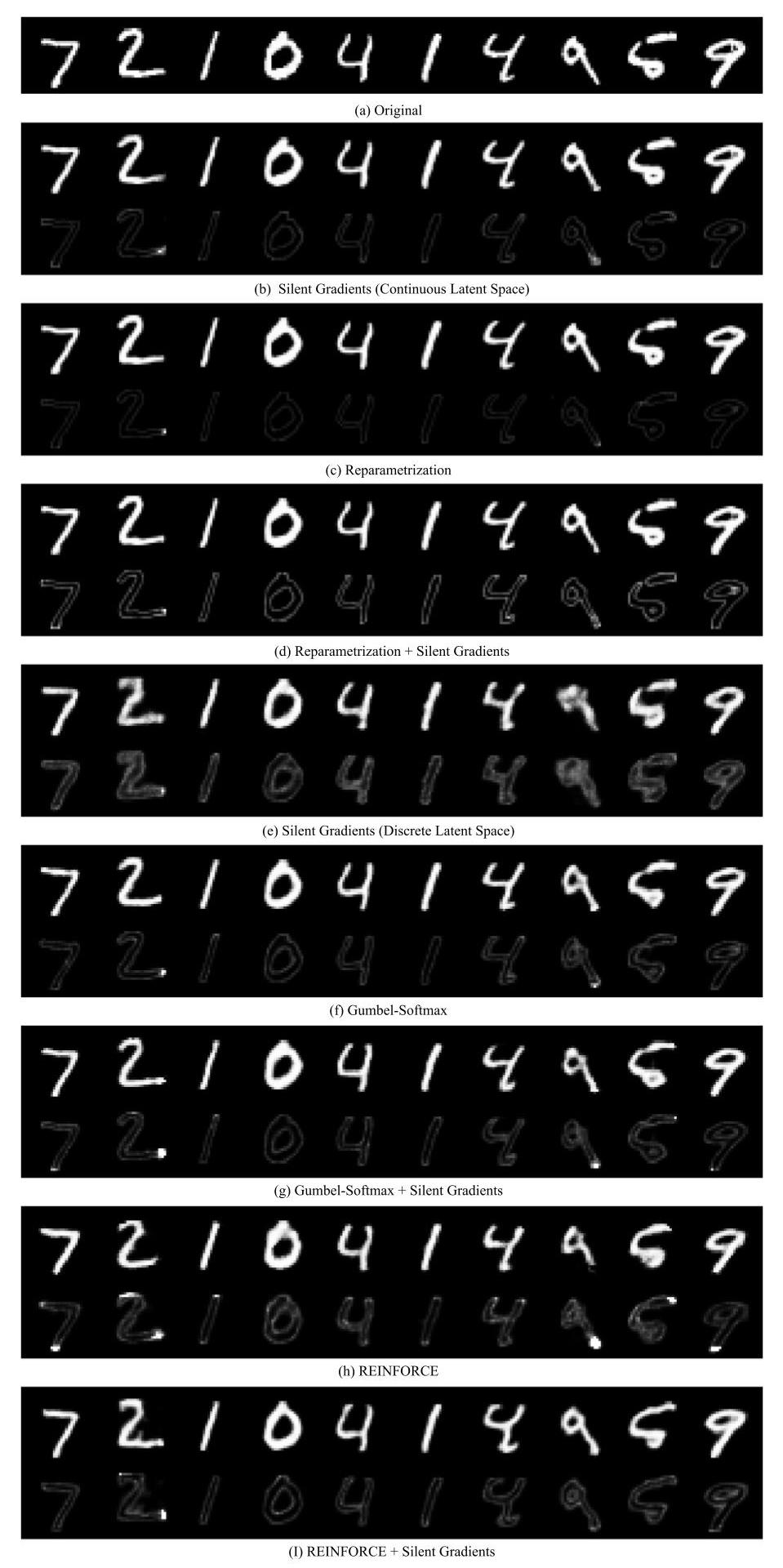}
        \caption{Reconstructions on the MNIST dataset in learnable variance experiment. The images are the output of the nonlinear decoder.}
        \label{fig:mnist_learnable}
\end{figure}

\begin{figure}[ht]
    \centering
        \includegraphics[width=0.48\textwidth]{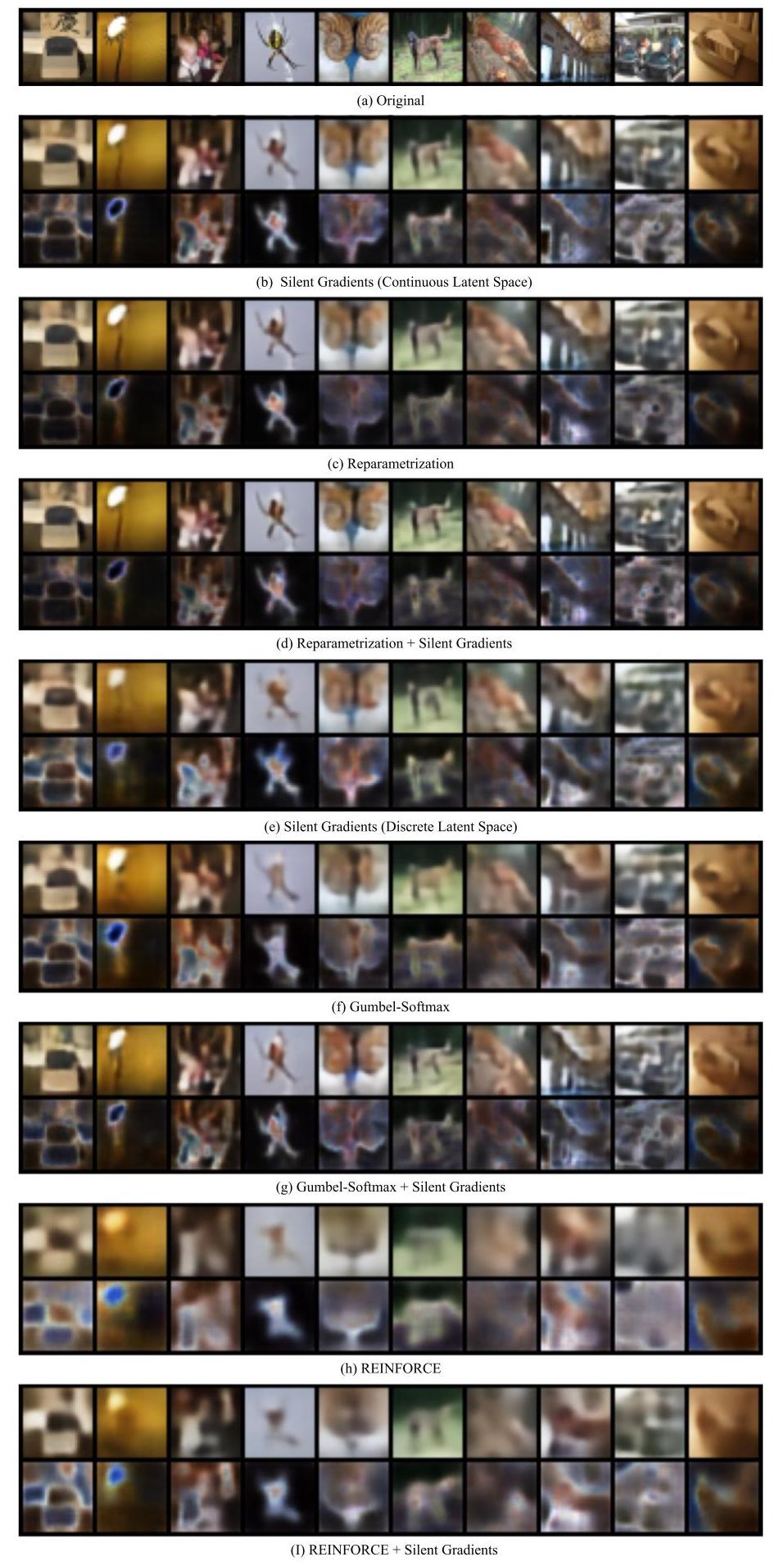}
        \caption{Reconstructions on the ImageNet dataset in learnable variance experiment.}
        \label{fig:imagenet}
\end{figure}

\begin{figure}[ht]
    \centering
        \includegraphics[width=0.48\textwidth]{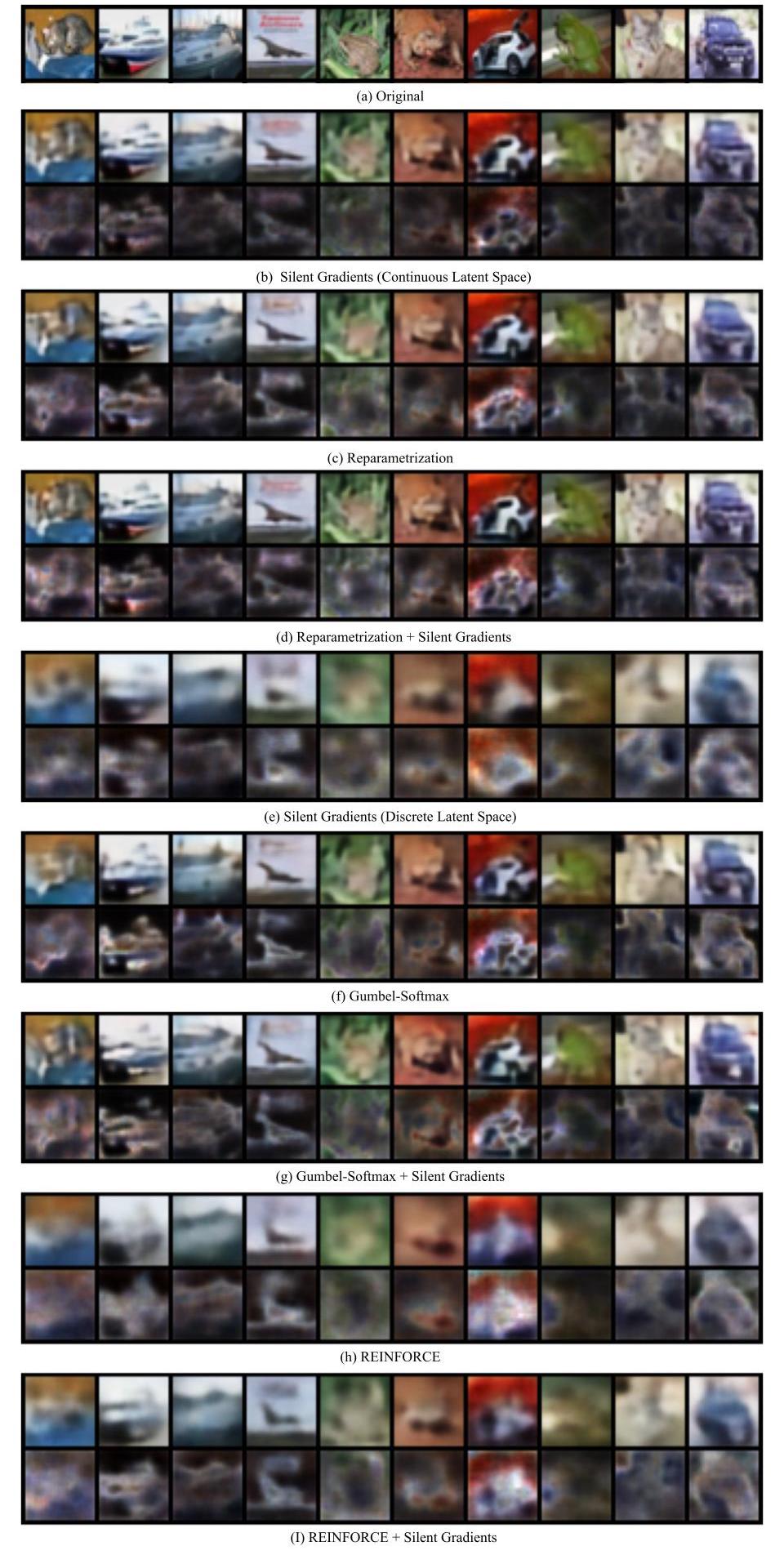}
        \caption{Reconstructions on the CIFAR-10 dataset in learnable variance experiment.}
        \label{fig:cifar10}
\end{figure}

In addition to the reconstructed mean, the visualizations include an additional row displaying the learned standard deviation for each pixel. For visualization purposes, the standard deviation is normalized to the range $[0, 1]$ to be displayed as an image, as shown in \Cref{fig:mnist_learnable}, \Cref{fig:imagenet}, and \Cref{fig:cifar10}.
    
\end{document}